\documentclass[acmsmall]{acmart}

\setcopyright{cc}
\setcctype{by}
\acmJournal{PACMMOD}
\acmYear{2026} \acmVolume{4} \acmNumber{1 (SIGMOD)} \acmArticle{56}
\acmMonth{2} \acmPrice{} \acmDOI{10.1145/3786670}

\received{July 2025}
\received[revised]{October 2025}
\received[accepted]{November 2025}

\usepackage{multirow}
\usepackage{threeparttable}
\usepackage{listings}
\usepackage{xcolor}
\usepackage{amsmath}

\usepackage{amssymb}
\usepackage{pifont}
\usepackage{algorithm}
\usepackage{algpseudocode}
\usepackage{booktabs}
\usepackage{graphicx}
\usepackage{siunitx}
\usepackage{subcaption}
\usepackage{geometry}
\usepackage{enumitem}
\usepackage{url}
\usepackage{adjustbox}

\newcommand{\cmark}{\ding{51}} % Check mark
\newcommand{\xmark}{\ding{55}} % X mark

\definecolor{codegreen}{rgb}{0,0.6,0}
\definecolor{codegray}{rgb}{0.5,0.5,0.5}
\definecolor{codepurple}{rgb}{0.58,0,0.82}
\definecolor{backcolour}{rgb}{0.95,0.95,0.92}
\definecolor{algokeyword}{rgb}{0.13, 0.13, 1}
\definecolor{algofunc}{rgb}{0.73, 0.13, 0.13}
\definecolor{algocomment}{rgb}{0, 0.5, 0}
\definecolor{errorred}{rgb}{0.7,0,0}
\definecolor{errorback}{rgb}{1,0.95,0.95}

\lstdefinestyle{mystyle}{
    backgroundcolor=\color{backcolour},   
    commentstyle=\color{codegreen},
    keywordstyle=\color{magenta},
    numberstyle=\tiny\color{codegray},
    stringstyle=\color{codepurple},
    basicstyle=\ttfamily\footnotesize,
    breakatwhitespace=false,         
    breaklines=true,                 
    captionpos=b,                    
    keepspaces=true,                 
    numbers=left,                    
    numbersep=5pt,                  
    showspaces=false,                
    showstringspaces=false,
    showtabs=false,                  
    tabsize=2
}

\lstdefinestyle{errorstyle}{
    backgroundcolor=\color{errorback},
    basicstyle=\ttfamily\scriptsize\color{errorred},
    breaklines=true,
    numbers=none,
    frame=none,
    breakatwhitespace=false,
    showspaces=false,
    showstringspaces=false,
    showtabs=false
}

\lstdefinelanguage{Markdown}{
    keywords={},
    keywordstyle=\color{blue},
    comment=[l]{\#},
    commentstyle=\color{codegreen},
    string=[s]{"}{"},
    stringstyle=\color{codepurple},
    morestring=[s]{`}{`},
    basicstyle=\ttfamily\footnotesize,
    breaklines=true,
    showstringspaces=false,
    frame=none,
    numbers=none,
    tabsize=2,
    captionpos=b
}

\begin{document}

\title[MultiVis-Agent: A Multi-Agent Framework with Logic Rules for Reliable...]{MultiVis-Agent: A Multi-Agent Framework with Logic Rules for Reliable and Comprehensive Cross-Modal Data Visualization}

\author{Jinwei Lu}
\email{jwlu18@gmail.com}
\orcid{0009-0000-0561-3277}
\affiliation{%
  \institution{The Hong Kong Polytechnic University}
  \city{Hong Kong SAR}
  \country{China}
}

\author{Yuanfeng Song}
\email{songyf@outlook.com}
\orcid{0000-0003-2221-9807}
\affiliation{%
  \city{Shanghai}
  \institution{ByteDance}
  \country{China}
}

\author{Chen Zhang}
\email{jason-c.zhang@polyu.edu.hk}
\orcid{0000-0002-3306-9317}
\affiliation{%
  \institution{The Hong Kong Polytechnic University}
  \city{Hong Kong SAR}
  \country{China}
}

\author{Raymond Chi-Wing Wong}
\email{raywong@cse.ust.hk}
\orcid{0000-0001-7045-6503}
\affiliation{%
  \institution{The Hong Kong University of Science and Technology}
  \city{Hong Kong SAR}
  \country{China}
}

\begin{abstract}
Real-world visualization tasks involve complex, multi-modal requirements that extend beyond simple text-to-chart generation, requiring reference images, code examples, and iterative refinement. Current systems exhibit fundamental limitations: single-modality input, one-shot generation, and rigid workflows. While LLM-based approaches show potential for these complex requirements, they introduce reliability challenges including catastrophic failures and infinite loop susceptibility. To address this gap, we propose \textbf{MultiVis-Agent}, a logic rule-enhanced multi-agent framework for reliable multi-modal and multi-scenario visualization generation. Our approach introduces a four-layer logic rule framework that provides mathematical guarantees for system reliability while maintaining flexibility. Unlike traditional rule-based systems, our logic rules are mathematical constraints that guide LLM reasoning rather than replacing it. We formalize the \textbf{MultiVis} task spanning four scenarios from basic generation to iterative refinement, and develop \textbf{MultiVis-Bench}, a benchmark with over 1,000 cases for multi-modal visualization evaluation. Extensive experiments demonstrate that our approach achieves 75.63\% visualization score on challenging tasks, significantly outperforming baselines (57.54-62.79\%), with task completion rates of 99.58\% and code execution success rates of 94.56\% (vs. 74.48\% and 65.10\% without logic rules), successfully addressing both complexity and reliability challenges in automated visualization generation.
\end{abstract}




\maketitle

\section{Introduction}
\label{sec:intro}

Despite extensive research in automated visualization systems within the database and data mining community \cite{10.14778/3681954.3681992, 10.14778/3494124.3494151, 10.14778/3025111.3025126,10.14778/3055330.3055333,10.1145/3534678.3539330,10.1145/3448016.3457261,10.1145/3448016.3452764}, existing approaches face fundamental limitations that prevent practical deployment. Current database-centric systems focus on single-modal text-to-visualization translation, lacking multi-modal capabilities—the ability to process diverse inputs including reference images, reference code, and existing visualizations beyond natural language—and iterative refinement mechanisms essential for real-world analytical workflows. These systems suffer from critical reliability issues: LLM-based agents produce inconsistent outputs, while traditional pipelines are too rigid for complex scenarios involving reference images, code, or iterative feedback.

\begin{figure}[t]
    \centering
    \includegraphics[width=\linewidth]{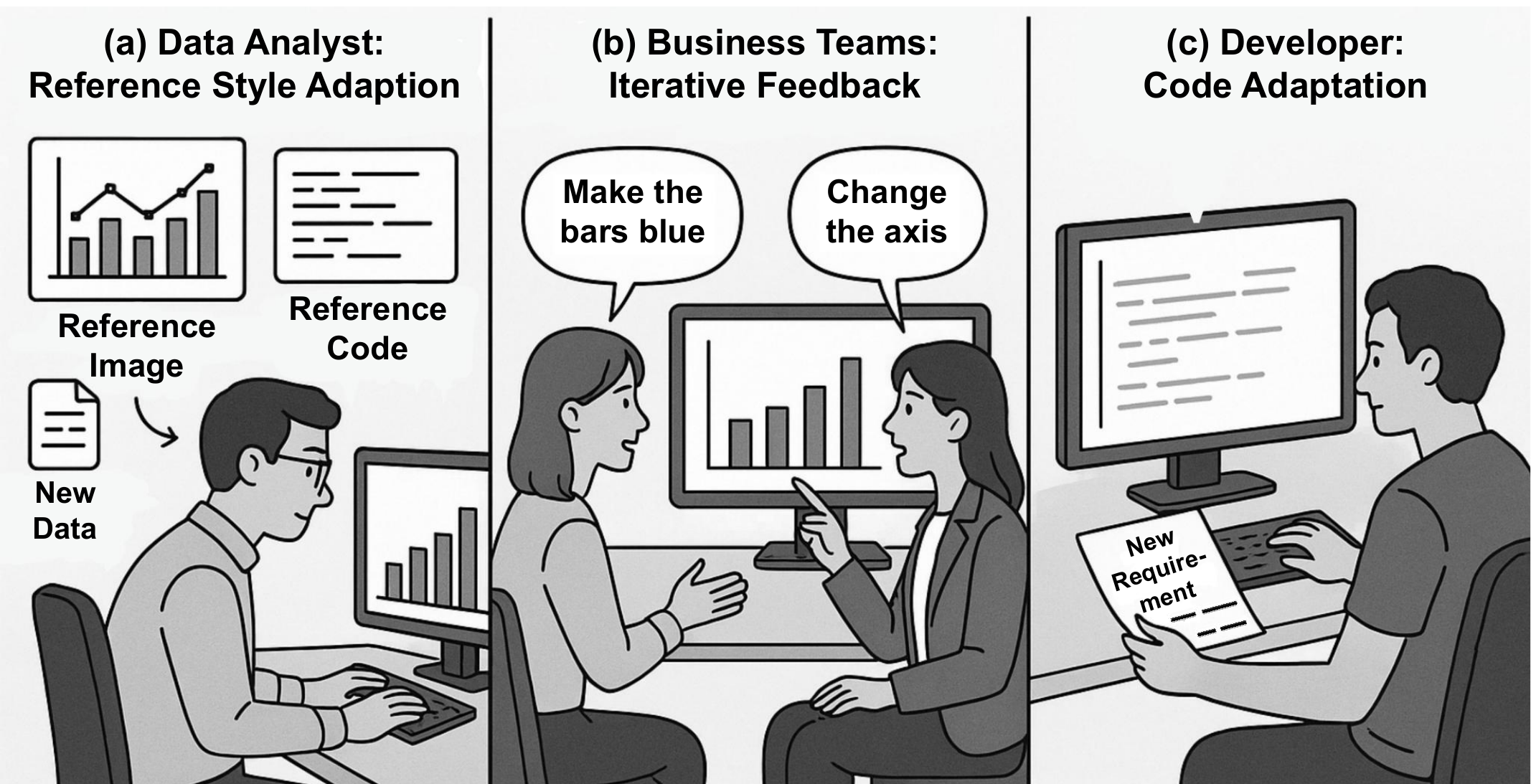}
    \caption{Real-world visualization tasks require multi-modal inputs and iterative refinement. Current Text-to-Vis systems fail to support these scenarios.}
    \label{fig:user_requirements}
\end{figure}

\paragraph{\textbf{System Limitations.}} Real-world visualization tasks extend far beyond simple text-to-vis generation (Figure~\ref{fig:user_requirements}). Analysts must adapt reference styles to new datasets, teams require iterative refinement based on feedback, and developers need to modify existing code for new requirements. Despite progress in Text-to-Visualization research \cite{10.1145/3448016.3457261,10121440,cheng-etal-2023-gpt,10443572,9617561,10.1145/3613904.3642943,9912366,wu-etal-2024-chartinsights,10530359, 10670526,10.1145/3534678.3539330,10.1145/3637528.3671935,10.1007/s00778-025-00912-0,10.1007/s00778-025-00954-4,11112835,11112929}, current systems exhibit three fundamental limitations: (1) \textbf{single-modality input}—systems like Chat2VIS \cite{10121440} cannot interpret reference images or code examples; (2) \textbf{one-shot generation}—approaches like Prompt4Vis \cite{10.1007/s00778-025-00912-0} employ single-pass processes unsuitable for iterative refinement; and (3) \textbf{rigid workflows}—systems like nvAgent \cite{ouyang2025nvagentautomateddatavisualization} use fixed processes that cannot adapt to diverse task types.

\paragraph{\textbf{The Reliability Crisis.}} LLM-based agent systems have introduced critical reliability challenges: (1) \textbf{catastrophic failures} with infinite error loops; (2) \textbf{uncontrolled error propagation} through entire workflows; (3) \textbf{parameter boundary violations} causing execution failures; and (4) \textbf{infinite loop susceptibility} in iterative processes lacking termination guarantees.

\paragraph{\textbf{Our Solution: MultiVis-Agent.}} To address these challenges, we propose \textbf{MultiVis-Agent}, a novel logic rule-enhanced multi-agent framework for reliable multi-modal visualization generation. Our approach decomposes complex visualization tasks into specialized sub-problems handled by coordinated agents. The centralized architecture enables consistent state management, global context awareness, and effective integration of diverse inputs. At its core, a \textbf{Coordinator Agent} dynamically orchestrates specialized agents for database interaction, code generation, and validation based on task context, enabling effective multi-modal fusion and state-aware iterative refinement.

\paragraph{\textbf{Key Innovations.}} We introduce a four-layer logic rule framework that provides mathematical guarantees for system reliability: (1) parameter boundary constraints, (2) deterministic task classification, (3) systematic error recovery, and (4) guaranteed loop termination. These formal constraints guide and constrain LLM-driven decisions while maintaining flexibility for complex visualization tasks.

\paragraph{\textbf{Contributions.}} Our work has four main contributions:

\begin{itemize}[leftmargin=*]\itemsep0em 
    \item \textbf{Logic Rule-Enhanced Multi-Agent Architecture:} We propose MultiVis-Agent \footnote{The source code and data are available at \url{https://github.com/Jinwei-Lu/MultiVis.git}.}, a novel framework with a four-layer logic rule system providing mathematical constraints for parameter safety, error recovery, and termination guarantees. This achieves task completion rates of 98.68-100\% and code execution success rates of 94.56-97.10\%, dramatically outperforming the same framework without logic rules (74.48-92.03\% and 63.18-83.33\% respectively).
    
    \item \textbf{MultiVis Task Formulation:} We formalize MultiVis, extending traditional Text-to-Vis to four scenarios: Basic Generation, Image-Referenced Generation, Code-Referenced Generation, and Iterative Refinement, systematically capturing real-world multi-modal visualization requirements.
    
    \item \textbf{MultiVis-Bench Benchmark:} We construct a novel benchmark with over 1,000 cases supporting multi-modal inputs and executable Python code output, enabling end-to-end evaluation unlike existing datasets with single-modal inputs and intermediate representations.
    
    \item \textbf{Superior Empirical Performance:} Extensive experiments show substantial improvements, with MultiVis-Agent achieving 75.63\% visualization quality in challenging Image-Referenced Generation versus 62.79\% (LLM Workflow) and 57.54\% (Instructing LLM). Performance remains strong across all scenarios (72.99-76.58\%), with logic rules contributing 17.58-31.70pp improvements.
\end{itemize}

The rest of this paper is organized as follows: Section~\ref{sec:multivis} introduces the MultiVis task formulation. Section~\ref{sec:multivis-bench} describes the MultiVis-Bench dataset. Section~\ref{sec:multivis-agent-framework} details the MultiVis-Agent framework. Section~\ref{sec:evaluation-metrics} explains our evaluation methodology. Section~\ref{sec:experimental-results} presents the experimental results. Section~\ref{sec:related-work} reviews related work, and Section~\ref{sec:conclusion} concludes the paper.

\section{MultiVis: A Comprehensive Task for Visualization Generation}
\label{sec:multivis}

\begin{table*}[t]
  \centering
  \caption{Comparison of MultiVis-Bench with existing visualization benchmarks.}
  \label{table:dataset_comparison}
  \resizebox{\textwidth}{!}{%
    \begin{threeparttable}
      \begin{tabular}{lccccccc}
          \toprule
          \multirow{2}{*}{\textbf{Datasets}} & 
          \multirow{2}{*}{\textbf{\#-Tables}} &
          \multirow{2}{*}{\textbf{\#-Samples}} &
          \multirow{2}{*}{\textbf{\begin{tabular}[c]{@{}c@{}}Ref. Image\\Input\end{tabular}}} &
          \multirow{2}{*}{\textbf{\begin{tabular}[c]{@{}c@{}}Ref. Code\\Input\end{tabular}}} &
          \multirow{2}{*}{\textbf{\begin{tabular}[c]{@{}c@{}}Output\\Format\end{tabular}}} &
          \multirow{2}{*}{\textbf{\begin{tabular}[c]{@{}c@{}}Scenario\\Types\tnote{*}\end{tabular}}} &
          \multirow{2}{*}{\textbf{\begin{tabular}[c]{@{}c@{}}Construction\\Leader\tnote{**}\end{tabular}}} \\
          & & & & & & & \\
          \midrule
          Quda~\cite{fu2020qudanaturallanguagequeries} & 36 & 14035 & \xmark & \xmark & Analytic Tasks & ATR & Human \\
          NLV Corpus~\cite{10.1145/3411764.3445400} & 3 & 814 & \xmark & \xmark & Vega-Lite & BG & Human \\
          Dial-NVBench~\cite{10.1145/3637528.3671935} & 780 & 124,449 & \xmark & \xmark & VQL & BG & Program \\
          VL2NL~\cite{10.1145/3613904.3642943} & 1,981 & 3,962 & \xmark & \xmark & Vega-Lite & BG & LLM \\
          VisEval~\cite{10670425} & 748 & 2,524 & \xmark & \xmark & VQL & BG & LLM \\
          nvBench~\cite{10.1145/3448016.3457261} & 780 & 25,750 & \xmark & \xmark & VQL & BG & Program \\
          nvBench 2.0~\cite{luo2025nvbench20benchmarknatural} & 780 & 24,076 & \xmark & \xmark & VQL & BG & LLM \\
          \midrule
          \textbf{MultiVis-Bench} & 697 & 1,202 & \cmark & \cmark & Python Code & BG, IRG, CRG, IR & Human+LLM \\
          \bottomrule
      \end{tabular}
      \begin{tablenotes}
        \small
        \item[*] \textbf{Scenario Types:} ATR: Analytic Task Recognition; BG: Basic Generation; IRG: Image-Referenced Generation; CRG: Code-Referenced Generation; IR: Iterative Refinement.
        \item[**] \textbf{Construction Leader:} Human: Expert-authored; Program: logic rule-based generation; LLM: AI-generated with minimal oversight; Human+LLM: Human-led with AI assistance and expert validation.
      \end{tablenotes}
    \end{threeparttable}%
  }
\end{table*}

\subsection{Overview and Motivation}

The prevailing Text-to-Visualization (Text-to-Vis) paradigm, while foundational, often struggles with the complexity and dynamism inherent in practical data analysis workflows. Its typical focus on single-turn, text-only requests does not fully capture the multi-modal nature of real-world inputs (e.g., reference images and code examples) or the essential iterative refinement process common in visualization design. To systematically address these challenges and establish a more encompassing framework, we propose \textbf{MultiVis}, a comprehensive task formulation that significantly extends traditional Text-to-Vis by unifying diverse generation and refinement scenarios.

The MultiVis task formulation offers several key advantages: (i) it explicitly supports multimodal inputs, reflecting real user requirements; (ii) it systematically covers key stages of the visualization \textit{generation and refinement} lifecycle; (iii) its structured categorization provides a clear framework for addressing diverse scenarios; and (iv) its progression from basic to complex situations facilitates systematic system development and evaluation. This comprehensive formulation motivates the need for flexible and adaptive system architectures, such as the agent-based approach detailed in Section~\ref{sec:multivis-agent-framework}, capable of handling the distinct requirements of each MultiVis scenario.

\subsection{Task Formulation}

Formally, we define MultiVis as a family of tasks aimed at learning functions that transform various combinations of inputs into executable visualization code. Let \( D \) represent a database, \( Q \) a natural language query (NLQ), \( I_{ref} \) a reference image, \( C_{ref} \) reference code, \( V_{old} \) existing visualization code to be refined, and \( V \) the target visualization code (either newly generated or refined). The core task within each MultiVis scenario is to learn a mapping \( f: \mathcal{X} \rightarrow V \), where \( \mathcal{X} \) is a specific combination of input modalities (such as \textit{Q}, \textit{D}, \textit{I\textsubscript{ref}}, \textit{C\textsubscript{ref}} and \textit{V\textsubscript{old}}). Based on the composition of these inputs and the specific goal, we categorize MultiVis into four distinct scenarios:

\begin{itemize}[leftmargin=*]\itemsep0em 
    \item \textbf{Scenario A (BG, Basic Generation):} Inputs \( \mathcal{X} = (Q, D) \); Output target is \( V \). This corresponds to traditional Text-to-Vis. \textit{Example:} $Q$: ``Show sales trends by month''; $D$: Table [month, sales\_amount] $\rightarrow$ $V$: \texttt{alt.Chart(data).mark\_line().encode(x='month:T', y='sales\_amount: Q')}. \textit{Existing systems:} Traditional approaches handle straightforward scenarios but struggle with ambiguous queries.

    \item \textbf{Scenario B (IRG, Image-Referenced Generation):} Inputs \( \mathcal{X} = (Q, D, I_{ref}) \); Output target is \( V \). The image \( I_{ref} \) provides style, layout, or chart type cues, introducing \textit{cross-modal understanding} challenges. \textit{Example:} $Q$: ``Create a visualization like this reference image''; $D$: Table [sales, category, region]; $I_{ref}$: Stacked bar chart with custom colors $\rightarrow$ $V$: \texttt{alt.Chart(data).mark\_bar().encode (x='sum(sales):Q', y='category:N', color='region:N')}. \textit{Existing systems:} Text-only approaches cannot interpret visual style cues, leading to generic visualizations.

    \item \textbf{Scenario C (CRG, Code-Referenced Generation):} Inputs \( \mathcal{X} = (Q, D, C_{ref}) \); Output target is \( V \). Reference code \( C_{ref} \) provides implementation patterns, posing \textit{code understanding and adaptation} challenges. \textit{Example:} $Q$: ``Adapt this matplotlib code using Altair and current data''; $D$: Table [category]; $C_{ref}$: \texttt{plt.scatter(x, y, c=colors, alpha=0.6)} $\rightarrow$ $V$: \texttt{alt.Chart(data).mark\_circle (opacity=0.6).encode(x= 'x:Q',y='y:Q',color='category: N')}. \textit{Existing systems:} Current approaches struggle with cross-library translation and semantic code understanding.

    \item \textbf{Scenario D (IR, Iterative Refinement):} Inputs \( \mathcal{X} = (Q, D, V_{old}) \); Output target is \( V \) (refined code). The existing code \( V_{old} \) is modified based on feedback, requiring \textit{dialogue state tracking and precise modification}. \textit{Example:} $Q$: ``Make x-axis labels vertical and add title''; $D$: Table [month]; $V_{old}$: \texttt{alt.Chart(data).mark\_bar(). encode(x='month:N', y='sales:Q')} $\rightarrow$ $V$: \texttt{...encode(x=alt. X('month:N', axis=alt.Axis(labelAngle=-90)), y='sales: Q').properties(title='Sales by Month')}. \textit{Existing systems:} One-shot approaches cannot handle iterative feedback loops.
\end{itemize}
For each scenario \( S \in \{A, B, C, D\} \), this mapping is learned by a specialized function \( f_S(\cdot) \in \mathcal{F}_{\theta_S} \), parameterized by \( \theta_S \). This explicit distinction underscores that effectively handling the unique input combinations and inherent challenges of each scenario (e.g., interpreting image styles, adapting code structures, and implementing precise modifications) likely requires different underlying model capabilities or architectural components, moving beyond a one-size-fits-all approach.

\section{MultiVis-Bench: Benchmark of MultiVis}
\label{sec:multivis-bench}

\begin{figure*}[htbp]
    \centering
    \includegraphics[width=\linewidth]{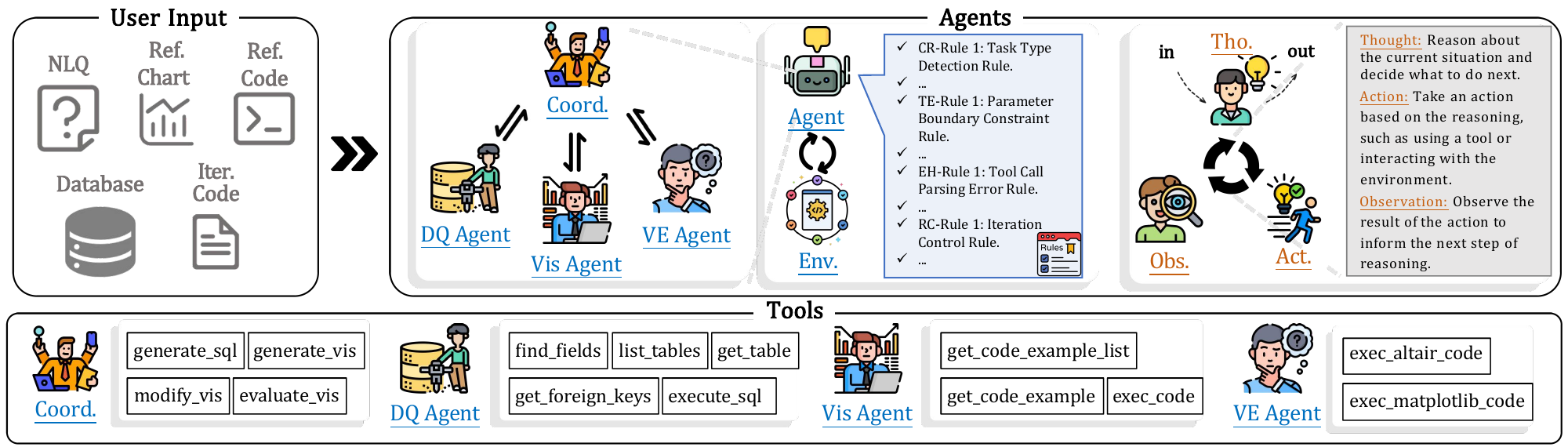}
    \caption{The architecture of MultiVis-Agent. The framework processes multi-modal user inputs (NLQ, Reference Chart, Reference Code, Iterative Code, Database) through a central Coordinator Agent that dynamically orchestrates specialized agents: the Database \& Query Agent (DQ Agent), the Visualization Implementation Agent (Vis Agent), and the Validation \& Evaluation Agent (VE Agent). Each agent is equipped with specific tools and follows a reasoning cycle (Thought-Observation-Action). The system is governed by a four-layer logic rule framework (CR, TE, EH, RC logic rules) that ensures reliable execution and error handling, enabling task-adaptive visualization generation and iterative refinement.}
    \label{fig:multivis-agent-architecture}
\end{figure*}

\subsection{Overview and Motivation}

To evaluate systems designed for complex, multi-modal, and iterative visualization tasks, which existing benchmarks often inadequately cover, we introduce \textbf{MultiVis-Bench}. It is a novel benchmark specifically constructed to assess capabilities across the comprehensive MultiVis scenarios (A-D) defined in Section~\ref{sec:multivis}. MultiVis-Bench focuses on crucial capabilities such as handling diverse \textbf{multi-modal inputs} (images, code), supporting \textbf{iterative refinement} processes, and generating directly \textbf{executable visualization code}. These aspects are critical for advanced systems like MultiVis-Agent but are poorly covered by benchmarks limited to basic Text-to-Vis or intermediate, non-executable representations.

Table~\ref{table:dataset_comparison} compares MultiVis-Bench with existing visualization benchmarks, contextualizing its contributions and highlighting its unique features, including multi-modal input support, executable Python code output, and full lifecycle scenario coverage.

\subsection{Dataset Construction Methodology}

MultiVis-Bench was constructed through a systematic, human-led approach with LLM assistance, focusing on quality, diversity, and representativeness across the four MultiVis scenarios.

\paragraph{Foundational Resources.} We prepared two essential resource types: (1) \textit{Databases}: 141 complex SQLite databases (average 5.8 tables, 27 columns) selected from the Spider benchmark \cite{yu-etal-2018-spider}, filtered for visualization suitability based on predefined criteria (presence of numerical/categorical data, sufficient data volume and domain diversity) across multiple domains including business, science, and social studies; (2) \textit{Visualization Templates}: Standardized Altair code templates for 127 distinct chart types (bar charts, scatter plots, area charts, composite charts, etc.) sourced from official Altair documentation and curated visualization examples.

\paragraph{Scenario-Specific Construction.} For each MultiVis scenario, we employed a human-led, LLM-assisted approach where initial candidates were generated by an LLM (Gemini-2.0-pro-exp) prompted with database schemas, visualization templates, and scenario-specific instructions, then underwent rigorous human expert review and refinement (averaging 2.5 rounds per example). 

For \textbf{Scenario A (BG)}, we provided Altair documentation and database to generate candidate pairs of (NLQ, visualization code), with human experts selecting or manually adjusting examples to ensure quality, resulting in 306 examples covering 127 chart types. \textbf{Scenario B (IRG)} involved human experts first curating 109 reference chart images (covering 107 chart types), then manually rephrasing and adapting queries from Scenario A to incorporate visual cues from reference images, instructing systems to generate visualizations matching reference styles with new data. \textbf{Scenario C (CRG)} utilized 132 Altair and 101 Matplotlib reference code snippets manually curated by experts, who adapted Scenario A queries to use these external examples as structural/stylistic guides, totaling 233 examples. \textbf{Scenario D (IR)} built upon Scenario A cases, where experts formulated follow-up modification instructions and crafted correctly modified code to simulate realistic iterative refinement processes, resulting in 554 examples.

\paragraph{Quality Assurance.} All examples underwent comprehensive validation ensuring: (1) \textbf{Technical Correctness} (executable, error-free code producing intended chart structures), (2) \textbf{Semantic Faithfulness} (accurate query-visualization alignment and reference adherence), and (3) \textbf{Perceptual Effectiveness} (clear, interpretable charts following visualization best practices). The human review process utilized detailed checklists covering code style, requirement adherence, data mapping accuracy, visual clarity, and chart type appropriateness. Additional automated quality control included: (i) code formatting standardization using tools like Black, and (ii) automated execution validation of all code snippets against their respective databases. Substandard examples were iteratively corrected or rejected if fundamental flaws persisted. The complete construction involved iterative LLM-assisted generation with rigorous human expert validation (averaging 2.5 review rounds per example), ensuring high-quality examples across all technical, semantic, and perceptual dimensions.

\subsection{Dataset Statistics and Characteristics}

MultiVis-Bench contains 1,202 carefully selected cases comprising four task types corresponding to the scenarios in Section~\ref{sec:multivis}: Scenario A (BG) with 306 examples; Scenario B (IRG) with 109 examples; Scenario C (CRG) with 233 examples; and Scenario D (IR) with 554 examples. The dataset covers 127 chart types and utilizes 141 complex SQLite databases, allowing targeted evaluation across different visualization stages with progressively complex input modalities beyond traditional text+database input.

As detailed in Table~\ref{table:dataset_comparison}, MultiVis-Bench's key distinctions include native support for \textbf{multi-modal inputs} (text, database, image, code), generating directly executable \textbf{Python code}, a hybrid \textbf{construction approach} (human-led with LLM assistance) balancing quality and scale, and comprehensive \textbf{scenario coverage} (Basic Generation to Iterative Refinement) for evaluating systems across the entire visualization lifecycle. Unlike previous benchmarks focusing on intermediate, non-executable representations (e.g., VQL and Vega-Lite) from text-only inputs, MultiVis-Bench produces directly executable Altair Python code, facilitating end-to-end evaluation and accommodating multi-modal inputs to reflect more realistic visualization workflows.

\section{MultiVis-Agent: System Implementation}
\label{sec:multivis-agent-framework}

This section details the design and implementation of \textbf{MultiVis-Agent}, our logic rule-enhanced multi-agent collaborative framework for multi-modal and multi-scenario visualization generation and refinement. We first provide an overview of our design philosophy and system architecture (Section~\ref{subsec:design_philosophy_overview}). We then illustrate the framework's operational workflow through a concrete execution example (Section~\ref{subsec:concrete_example}). Next, we present our core innovation: the logic rule-enhanced agent architecture that ensures reliable behavior while maintaining flexibility for complex visualization tasks (Section~\ref{subsec:rule_guided_architecture}). Finally, we describe the multi-agent coordination framework that orchestrates specialized agents for complex visualization tasks (Section~\ref{subsec:coordination_framework}).

\subsection{Design Philosophy and Architecture Overview}
\label{subsec:design_philosophy_overview}

Traditional visualization generation systems face fundamental limitations that hinder their practical deployment. Pure LLM-based agents exhibit inherent instability, producing inconsistent outputs and lacking robust error handling mechanisms. Traditional pipeline-based methods, while predictable, are too rigid for complex multi-modal scenarios. To address these challenges, we propose a novel approach that combines the reliability of logic rule-enhanced behavior with the flexibility of multi-agent coordination.

\begin{figure*}[t!]
    \centering
    \includegraphics[width=\linewidth]{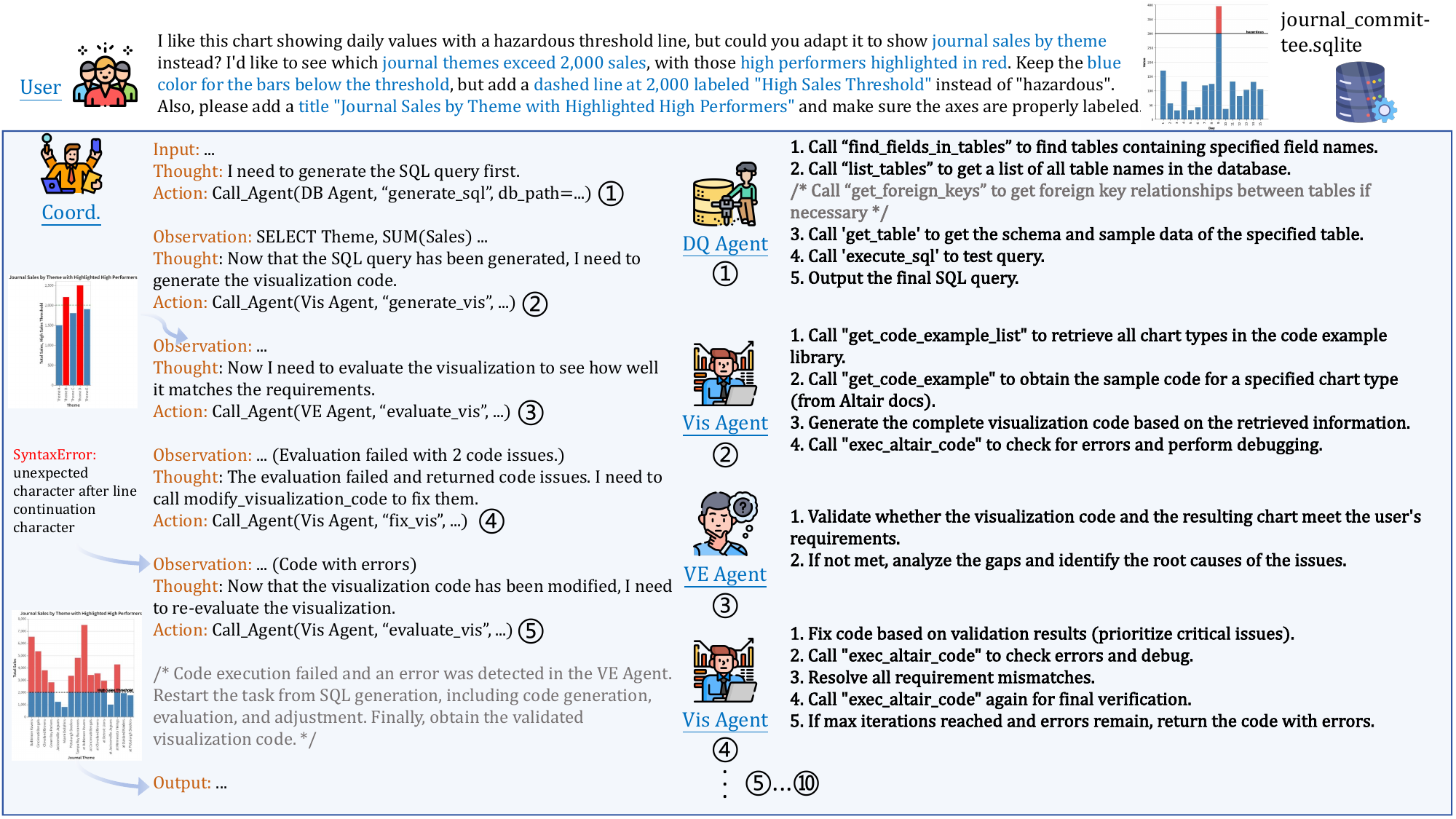}
    \caption{An example for the working process of MultiVis-Agent.}
    \label{fig:multivis-agent-example}
\end{figure*}

\subsubsection{Architecture Overview}

MultiVis-Agent employs a centralized multi-agent architecture that provides both specialization and flexibility. As depicted in Figure~\ref{fig:multivis-agent-architecture}, the framework decomposes complex visualization tasks into sub-problems handled by specialized agents (Database \& Query, Visualization Implementation, Validation \& Evaluation), while a central Coordinator manages the overall process through logic rule-enhanced decision making.

The centralized approach is particularly advantageous for visualization tasks because it enables: (1) maintaining global consistency across visual elements during generation and refinement, (2) integrating holistic feedback from validation processes, (3) facilitating robust state management and history tracking for effective iterative refinement, and (4) providing better error isolation and recovery mechanisms through our logic rule framework.

\subsection{Concrete Execution Example}
\label{subsec:concrete_example}

This subsection provides an overview of the technical implementation of MultiVis-Agent. Figure~\ref{fig:multivis-agent-example} illustrates the framework's operational workflow, showcasing how the Coordinator Agent orchestrates specialized agents to handle user requests through cycles of data retrieval, visualization generation, evaluation, and refinement. MultiVis-Agent employs three specialized agents, each with distinct responsibilities:

(i) \textit{Database \& Query Agent}. Understands data requirements and retrieves suitable data through five core tools: \texttt{list\_tables}, \texttt{get\_table}, \texttt{get\_foreign\_keys}, \texttt{find\_fields}, and \texttt{execute\_sql}. The main \texttt{generate\_sql} interface (as illustrated in Figure~\ref{fig:multivis-agent-example}\textcircled{1}) performs iterative exploration to formulate correct SQL queries.

(ii) \textit{Visualization Implementation Agent}. Translates data and requirements into executable Altair code using three tools: \texttt{get\_code\_ example\_list}, \texttt{get\_code\_example}, and \texttt{execute\_code}. Provides two interfaces: \texttt{generate\_visualization\_code} for initial generation (as illustrated in Figure~\ref{fig:multivis-agent-example}\textcircled{2}) and \texttt{modify\_visualization\_code} (demonstrated in Figure~\ref{fig:multivis-agent-example}\textcircled{4}) for iterative refinement.

(iii) \textit{Validation \& Evaluation Agent}. Assesses technical correctness and perceptual effectiveness using two tools: \texttt{execute\_altair \_code} and \texttt{execute\_matplotlib\_code}. The \texttt{evaluate\_visualiza tion} interface (illustrated in Figure~\ref{fig:multivis-agent-example}\textcircled{3} and \textcircled{5}) performs structured assessment and provides actionable feedback for refinement.

\subsection{Logic rule-enhanced Agent Architecture}
\label{subsec:rule_guided_architecture}

Our core innovation lies in the logic rule-enhanced agent architecture that ensures reliable behavior while maintaining flexibility for complex visualization tasks. In the formal definitions that follow, we use $\mathcal{F}_S(\cdot) \in \mathcal{F}_{\theta_S}$ to denote the learned mapping function for scenario $S$ parameterized by $\theta_S$, $\mathcal{T}_{tool}$ for the set of available tools, $\mathcal{R} = \{CR, TE, EH, RC\}$ for our four-layer logic rule framework, predicates like $\mathcal{V}(t,s)$ for validation functions, and $T_{max} = 10$ for the maximum iteration bound. We establish the design principles, introduce the four-layer logic rule framework, and provide theoretical guarantees.

\subsubsection{Logic rule-enhanced Design Principles}
\label{subsubsec:design_philosophy}

Our key insight is that \textbf{mathematical constraints} can guide and constrain LLM-driven decisions while preserving adaptability, fundamentally addressing the system robustness crisis. We introduce \textit{logic rule-enhanced agents}—systems where agent behavior is constrained by formally defined logic rules that ensure graceful error recovery and prevent system crashes. Unlike traditional rule-based systems that rely on static if-then rules, our logic rules are mathematically formalized constraints that provide formal guarantees while maintaining the flexibility of LLM reasoning.

Our logic rule-enhanced architecture is built on four fundamental principles: (1) \textbf{System Robustness}: Prevent catastrophic failures and enable graceful degradation under error conditions; (2) \textbf{Error Recovery Mechanisms}: Implement structured error detection, classification, and targeted recovery strategies; (3) \textbf{Execution Safety}: Ensure safe tool invocations through parameter validation and boundary enforcement; and (4) \textbf{Termination Guarante}: Provide mathematical assurance of finite execution time and loop prevention.

\subsubsection{Four-Layer logic rule Framework}
\label{subsubsec:rule_framework}

Our logic rule framework consists of four hierarchical layers that define explicit behavioral constraints through formal mathematical specifications. These layers are applied hierarchically across our agent architecture: CR rules govern the Coordinator Agent's task classification and decision logic; TE and EH rules constrain all four agents' tool execution and error handling; RC rules manage the entire system's iteration control and termination.

\paragraph{Coordination logic rules (CR): System-Level Decision Making.} 

\subparagraph{\textbf{(i) CR-Rule 1}: Task Type Detection.} Define task classification function $\mathcal{F}_T: \mathcal{I} \rightarrow \mathcal{T}_{task}$ that maps input domain $\mathcal{I}$ to task types $\mathcal{T}_{task} = \{A, B, C, D\}$ with priority ordering $D \succ C \succ B \succ A$:
\begin{equation}
\mathcal{F}_T(I) = \arg\max_{t \in \mathcal{T}_{task}} \pi(t, I)
\end{equation}
where $I \in \mathcal{I}$ is an input instance, $t \in \mathcal{T}_{task}$ is a task type, and $\pi: \mathcal{T}_{task} \times \mathcal{I} \rightarrow \mathbb{R}$ is the priority function. This ensures deterministic task identification for existing code (D), Python reference (C), image reference (B), or basic text (A).

The priority ordering $D \succ C \succ B \succ A$ distinguishes input explicitness to ensure complete task coverage: existing code (D) enables precise modification; code references (C) provide explicit structural guidance; image references (B) require expensive cross-modal understanding; basic text (A) starts from scratch. When multiple input types coexist (e.g., both image and code), the system classifies the task by the highest-priority category, guaranteeing deterministic routing and optimal resource utilization.

\subparagraph{\textbf{(ii) CR-Rule 2}: Tool Prerequisite Validation.} Let $\mathcal{T}_{tool}$ denote the set of available tools and $\mathcal{S}$ the set of all possible system states. Define validation predicate $\mathcal{V}: \mathcal{T}_{tool} \times \mathcal{S} \rightarrow \{0,1\}$ that ensures tool execution safety:
\begin{equation}
\mathcal{V}(t, s) = \mathbf{1}[\mathcal{P}(t) \subseteq \mathcal{D}(s)]
\end{equation}
where $t$ is a tool, $s$ is system state, $\mathcal{P}(t)$ denotes prerequisite set for tool $t$, $\mathcal{D}(s)$ denotes defined attributes in state $s$, and $\mathbf{1}[\cdot]$ is the indicator function.

\subparagraph{\textbf{(iii) CR-Rule 3}: Evaluation Response Decision.} Let $\mathcal{E}_{result}$ be the set of possible evaluation results and $\mathcal{A}_{action}$ the set of possible system actions. Define decision function $\mathcal{N}: \mathcal{E}_{result} \rightarrow \mathcal{A}_{action}$ that maps evaluation results to system actions:
\begin{equation}
\mathcal{N}(e) = f_d(e.passed, |e.recommendations|)
\end{equation}
where $e \in \mathcal{E}_{result}$ is evaluation result, $f_d$ is deterministic mapping function, $e.passed$ indicates success, and $|e.recommendations|$ is recommendation count.

For example, if validation passes without recommendations, the decision function returns "task complete." If validation fails with some critical issues, it returns "modify with prioritized feedback." This deterministic mapping ensures consistent system behavior.

\paragraph{Tool Execution logic rules (TE): Parameter Safety and Environment Control.}

\subparagraph{\textbf{(i) TE-Rule 1}: Parameter Boundary Constraint.} Let $\mathcal{P}$ denote the set of all tool parameters. Define constraint function $\phi: \mathcal{P} \times \mathbb{R} \rightarrow \mathbb{R} \cup \{\bot\}$ that enforces safety bounds:
\begin{equation}
\phi(p, v) = \begin{cases}
\max(\min(v, u_p), l_p) & \text{if } v \in \mathcal{V}_p \\
\bot & \text{otherwise}
\end{cases}
\end{equation}
where $p \in \mathcal{P}$ is parameter, $v$ is its value, $[l_p, u_p]$ is valid range, $\mathcal{V}_p$ is validity domain, and $\bot$ denotes invalid outcome.

\subparagraph{\textbf{(ii) TE-Rule 2}: Code Execution Environment.} Let $\mathcal{C}$ be the set of all possible input code snippets and $\mathcal{C}'$ the set of all possible standardized code snippets. Define standardization function $\psi: \mathcal{C} \rightarrow \mathcal{C}'$ that ensures execution consistency:
\begin{equation}
\psi(c) = \mathcal{N}(c) \cup \mathcal{M}(c) \cup \mathcal{I}(c)
\end{equation}
where $c \in \mathcal{C}$ is code snippet, $\mathcal{N}(c)$ performs namespace standardization, $\mathcal{M}(c)$ applies pattern matching, and $\mathcal{I}(c)$ injects save operations.

\subparagraph{\textbf{(iii) TE-Rule 3}: Reference Material Processing.} Let $\mathcal{R}_{type}$ be the set of reference material types, $\mathcal{F}_{path}$ the set of possible file paths, and $\mathcal{P}_{ref}$ the set of processed reference materials. Define processing function $\rho: \mathcal{R}_{type} \times \mathcal{F}_{path} \rightarrow \mathcal{P}_{ref}$ that enforces reference utilization:
\begin{equation}
\rho(t, f) = \mathcal{T}_{proc}(t) \circ \mathcal{V}_{file}(f) \circ \mathcal{P}_{proc}(t, f)
\end{equation}
where $t \in \mathcal{R}_{type}$ is reference type, $f \in \mathcal{F}_{path}$ is file path, $\mathcal{T}_{proc}(t)$ determines processing type, $\mathcal{V}_{file}(f)$ validates file accessibility, $\mathcal{P}_{proc}(t, f)$ applies type-specific processing, and $\circ$ denotes function composition.

\paragraph{Error Handling logic rules (EH): Systematic Error Recovery.} Let $\mathcal{E}_{error}$ denote the set of all possible errors, $\mathcal{C}_{context}$ the set of possible execution contexts, and $\mathcal{R}_{recovery}$ the set of possible recovery actions.

\subparagraph{\textbf{(i) EH-Rule 1}: Tool Call Parsing Error.} Let $\mathcal{T}_{text}$ be the set of all possible text outputs from LLMs and $\mathcal{E}_{type}$ the set of specific error types. Define error classification function $\epsilon: \mathcal{T}_{text} \rightarrow \mathcal{E}_{type}$:
\begin{equation}
\epsilon(x) = \mathcal{C}(\mathcal{S}(x), \mathcal{T}_{struct}(x), \mathcal{O}(x))
\end{equation}
where $x \in \mathcal{T}_{text}$ is input text, $\mathcal{S}(x)$, $\mathcal{T}_{struct}(x)$, $\mathcal{O}(x)$ analyze syntax, structure, and content respectively, and $\mathcal{C}$ performs classification.

\subparagraph{\textbf{(ii) EH-Rule 2}: Code Execution Error Recovery.} Define recovery function $\delta: \mathcal{E}_{error} \times \mathcal{C}_{context} \rightarrow \mathcal{R}_{recovery}$:
\begin{equation}
\delta(e, c) = \mathcal{R}_{\tau(e)}(e, c)
\end{equation}
where $e \in \mathcal{E}_{error}$ is error instance, $c \in \mathcal{C}_{context}$ is context, $\tau(e)$ determines error type.

\subparagraph{\textbf{(iii) EH-Rule 3}: Image Processing Validation.} Let $\mathcal{I}_{path}$ be the set of possible image paths. Define validation predicate $\nu: \mathcal{I}_{path} \rightarrow \{0,1\}$:
\begin{equation}
\nu(p) = \mathcal{F}(p) \land \mathcal{E}(p) \land \mathcal{C}(p)
\end{equation}
where $p \in \mathcal{I}_{path}$ is image path, $\mathcal{F}(p)$, $\mathcal{E}(p)$, $\mathcal{C}(p)$ validate format, existence, and encoding respectively, and $\land$ denotes the logical AND operator.

\paragraph{ReAct Control logic rules (RC): Iteration Management.} Let $\mathcal{R}_{response}$ be the set of possible agent responses.

\subparagraph{\textbf{(i) RC-Rule 1}: Iteration Control.} Define termination predicate $\tau: \mathbb{N} \times \mathcal{R}_{response} \rightarrow \{0,1\}$:
\begin{equation}
\tau(i, r) = \mathbf{1}[i \geq T_{max}] \lor \mathcal{F}(r) \lor \mathcal{L}(r)
\end{equation}
where $i$ is iteration count, $r \in \mathcal{R}_{response}$ is agent response, $T_{max}$ is maximum threshold, $\mathcal{F}(r)$ indicates final response, $\mathcal{L}(r)$ indicates error limit exceeded, and $\lor$ is the logical OR operator.

\subparagraph{\textbf{(ii) RC-Rule 2}: Response Format Validation.} Define validation function $\sigma: \mathcal{R}_{response} \rightarrow \{0,1\}$:
\begin{equation}
\sigma(r) = \mathcal{S}_{valid}(r) \land \mathcal{C}_{valid}(r)
\end{equation}
where $r \in \mathcal{R}_{response}$ is agent response, $\mathcal{S}_{valid}(r)$ validates structure, $\mathcal{C}_{valid}(r)$ validates content.

\subparagraph{\textbf{(iii) RC-Rule 3}: Model Selection.} Let $\mathcal{C}_{content}$ be the set of all possible input contents, $\mathcal{U}_{urls}$ the set of all possible URLs, $\mathcal{M}_{type}$ the set of available model types, and $\mathcal{M}$ the set of available models. Define selection function $\mu: \mathcal{C}_{content} \times \mathcal{U}_{urls} \rightarrow \mathcal{M}_{type}$:
\begin{equation}
\mu(c, u) = \arg\max_{m \in \mathcal{M}} \kappa(m, c, u)
\end{equation}
where $c \in \mathcal{C}_{content}$ is content, $u \in \mathcal{U}_{urls}$ is URL, $m \in \mathcal{M}$ is model, and $\kappa$ is compatibility function.

\subsubsection{Theoretical Guarantees}
\label{subsubsec:theoretical_guarantees}

Our logic rule-enhanced approach provides formal reliability guarantees through mathematically constrained agent behavior. Let $\mathcal{C} = (\mathcal{S}, \mathcal{A}, \mathcal{T}_{tool}, \mathcal{R})$ denote an execution context with system state set $\mathcal{S}$, agent action set $\mathcal{A}$, tool collection $\mathcal{T}_{tool}$, and logic rule framework $\mathcal{R} = \{CR, TE, EH, RC\}$.

\paragraph{Formal Theorems.} Our framework provides mathematical guarantees through four fundamental theorems:

\subparagraph{\textbf{(i) Theorem 1} (Parameter Safety).} \textit{For any execution context $\mathcal{C}$ consisting of system state set $\mathcal{S}$, agent action set $\mathcal{A}$, tool collection $\mathcal{T}_{tool}$, and logic rule framework $\mathcal{R} = \{CR, TE, EH, RC\}$, all tool executions satisfy parameter safety constraints:}
\begin{equation}
\forall t \in \mathcal{T}_{tool}, \forall \vec{p} \in \mathcal{P}^n: \mathcal{E}(t, \vec{p}) \Rightarrow \mathcal{S}(t, \vec{p}) \lor \mathcal{G}
\end{equation}
where $\mathcal{E}(t, \vec{p})$ denotes execution of tool $t$ with parameters $\vec{p}$, $\mathcal{S}(t, \vec{p})$ indicates safe execution, and $\mathcal{G}$ represents graceful failure.

\textbf{Proof Sketch:} Partition parameter domain $\mathcal{P}^n = \mathcal{P}_{valid}^n \cup \mathcal{P}_{invalid}^n$ where $\mathcal{P}_{valid}^n \cap \mathcal{P}_{invalid}^n = \emptyset$. For valid parameters $\vec{p} \in \mathcal{P}_{valid}^n$, TE-Rule 1 ensures parameter boundary constraint function $\phi(p_i, \text{bounds}) = p_i$ and CR-Rule 2 validates prerequisites via $\mathcal{V}(t, s) = 1$, guaranteeing safe execution $\mathcal{S}(t, \vec{p})$. For invalid parameters $\vec{p} \in \mathcal{P}_{invalid}^n$, TE-Rule 1 produces $\phi(p_i, \text{bounds}) = \bot$, triggering EH-Rules 1-3 for error classification $\epsilon(\text{``invalid parameter''}) = \mathcal{E}_{param}$, recovery strategy selection $\delta(\mathcal{E}_{param}, \mathcal{C}) = \mathcal{R}_{param}$, and graceful failure $\mathcal{G}$. Exhaustive case analysis covers all parameter vectors.

\subparagraph{\textbf{(ii) Theorem 2} (Bounded Error Recovery).} \textit{For any error $e \in \mathcal{E}_{error}$ encountered during system execution, there exists a finite bound $T_e \in \mathbb{N}$ such that the error recovery function $\delta: \mathcal{E}_{error} \times \mathcal{C} \rightarrow \mathcal{R}_{recovery}$ terminates within $T_e$ steps:}
\begin{equation}
\forall e \in \mathcal{E}_{error}: \exists T_e \in \mathbb{N}, \delta(e, \cdot) \text{ terminates within } T_e \text{ steps}
\end{equation}

\textbf{Proof Sketch:} Use strong mathematical induction on error recovery attempts. Define time bounds $T_{classify}^{max}, T_{strategy}^{max}(\tau), T_{validate}^{max} \\< \infty$ for each recovery component where $\tau = \tau_{error}(e) \in \{\mathcal{E}_{param},\\\mathcal{E}_{exec}, \mathcal{E}_{parse}, \mathcal{E}_{timeout}\}$. Base case: first attempt takes time $T_1 \leq T_{single}^{max} < \infty$. Inductive step: if $n$ attempts take time $T_n \leq n \cdot T_{single}^{max}$, then $(n+1)$-th attempt satisfies $T_{n+1} \leq (n+1) \cdot T_{single}^{max}$. RC-Rule 1 enforces finite retry limit $N_{max} \geq 1$, guaranteeing termination with bound $T_e = N_{max} \cdot T_{single}^{max} < \infty$.

\subparagraph{\textbf{(iii) Theorem 3} (Guaranteed Termination).} \textit{All execution sequences $\sigma \in \Sigma$ (where $\Sigma$ denotes the set of all possible execution traces) in the MultiVis-Agent framework terminate within a maximum of $T_{max} = 10$ iterations:}
\begin{equation}
\forall \sigma \in \Sigma: |\sigma| \leq T_{max}
\end{equation}
where $\sigma$ represents an execution trace and $|\sigma|$ denotes its length.

\textbf{Proof Sketch:} Use loop invariants and well-ordering principle of natural numbers. Define execution sequence $\sigma = \langle s_0, a_1, s_1, \ldots, a_k, s_k \rangle$ with state transition function $\mathcal{T}: \mathcal{S} \times \mathcal{A} \times \mathcal{R} \rightarrow \mathcal{S}$. Define loop invariant $I(i): \text{counter}(s_i) = i \land 0 \leq i \leq T_{max}$. Base case $I(0)$ holds with $\text{counter}(s_0) = 0$. Inductive step: either early termination occurs via $\text{task\_complete}(s_i) = 1$, $\text{fatal\_error}(s_i) = 1$, or $\text{resources\_exhausted}(s_i) = 1$, or $I(i+1)$ holds. At $i = T_{max}$, termination predicate $\tau(T_{max}, r) = \mathbf{1}[T_{max} \geq T_{max}] \lor \mathcal{F}(r) \lor \mathcal{L}(r) = 1$ forces mandatory termination. Contradiction argument eliminates sequences with $|\sigma| > T_{max}$.

\subparagraph{\textbf{(iv) Theorem 4} (System Reliability).} \textit{The MultiVis-Agent framework provides overall system reliability through the combination of parameter safety, error recoverability, and termination guarantees:}
\begin{equation}
\mathcal{R}(\mathcal{C}) \equiv \mathcal{P}(\mathcal{C}) \land \mathcal{E}(\mathcal{C}) \land \mathcal{T}_{term}(\mathcal{C})
\end{equation}
where $\mathcal{P}(\mathcal{C})$, $\mathcal{E}(\mathcal{C})$, and $\mathcal{T}_{term}(\mathcal{C})$ denote parameter safety, error recoverability, and termination guarantee respectively.

\textbf{Proof Sketch:} Define system reliability as logical conjunction $\mathcal{R}(\mathcal{C}) \triangleq \mathcal{P}(\mathcal{C}) \land \mathcal{E}(\mathcal{C}) \land \mathcal{T}_{term}(\mathcal{C})$ where $\mathcal{P}(\mathcal{C})$ denotes parameter safety from Theorem 1, $\mathcal{E}(\mathcal{C})$ denotes bounded error recovery from Theorem 2, and $\mathcal{T}_{term}(\mathcal{C})$ denotes guaranteed termination from Theorem 3. Component verification establishes $\mathcal{P}(\mathcal{C}) = \mathcal{E}(\mathcal{C}) = \mathcal{T}_{term}(\mathcal{C}) = \text{True}$. Logical composition yields $\mathcal{R}(\mathcal{C}) = \text{True} \land \text{True} \land \text{True} = \text{True}$. Necessity-sufficiency analysis proves these components constitute complete reliability definition for autonomous agent systems. Universal quantification extends over all execution contexts sharing logic rule framework $\mathcal{R} = \{CR, TE, EH, RC\}$. 

These theorems provide formal guarantees for system safety and termination, enhancing the reliability of our framework for practical deployment.

\paragraph{Implementation Robustness Considerations.} The reliability of our theoretical guarantees depends on robust constraint implementation through: (1) constraint-guided decision functions that bound LLM decisions within safe boundaries; (2) systematic validation mechanisms that verify prerequisites before tool execution; (3) efficient rule enforcement designed for rapid execution; and (4) structured fallback mechanisms ensuring graceful degradation. These implementations ensure mathematical guarantees hold in practice through deterministic threshold-based logic, prerequisite verification, and structured fallback mechanisms.

\paragraph{Formal logic rule Application Workflow.} Logic rules are applied hierarchically: CR rules first classify tasks and validate prerequisites; TE rules then enforce parameter safety and environment standardization; EH rules monitor and recover from errors during execution; finally RC rules ensure termination. This layered approach provides mathematical guarantees for system robustness while maintaining flexibility for complex multi-modal visualization tasks.

\paragraph{Clarification: logic rule-LLM Interaction Balance.} Our logic rule framework operates through a \textbf{constraint-guided paradigm} where LLM reasoning is preserved for content generation but constrained by mathematical logic rules for system behavior. Specifically: (1) \textbf{LLM Content Generation}: Specialized agents use LLM reasoning for generating SQL queries, visualization code, and evaluation assessments within their ``Thought-Action-Observation'' cycles; (2) \textbf{Logic Rule-Constrained Decisions}: All system-level decisions (task routing, tool selection, error recovery, termination) are governed by deterministic mathematical functions, not LLM reasoning; (3) \textbf{Validation and Enforcement}: logic rules act as validators and enforcers that override LLM outputs when they violate safety constraints or system policies; and (4) \textbf{Hybrid Reliability}: This approach combines the creativity of LLM content generation with the reliability of logic rule-based system control, ensuring both flexibility and mathematical guarantees.

For example, when the Database Agent generates a SQL query, CR-Rule 2 first validates prerequisites ($\mathcal{V}(t, s)$) before execution. If errors occur, EH-Rule 2 deterministically classifies the error type ($\epsilon$) and selects recovery strategies ($\delta$) without LLM involvement. The LLM then generates corrections based on structured feedback, but RC-Rule 1's termination predicate ($\tau$) mathematically enforces the $T_{max}=10$ bound. This separation ensures LLMs provide flexible content generation while safety-critical decisions follow provably correct mathematical constraints.

Unlike traditional agent systems that fail catastrophically when encountering errors, our logic rule-enhanced approach ensures graceful degradation and systematic recovery, making it suitable for production deployment.

\paragraph{Distinction from Traditional Rule-Based Systems.} It is crucial to distinguish our logic rule framework from traditional rule-based systems. While traditional rule-based approaches rely on static if-then rules that rigidly dictate system behavior, our logic rules are \textbf{mathematical constraints} that guide but do not replace LLM reasoning. Our approach preserves the generative flexibility of LLMs while providing formal guarantees through constraint-guided decision making, error recovery mechanisms, and termination conditions—a fundamentally different paradigm from rigid rule-based systems.

\subsection{Multi-Agent Coordination Framework}
\label{subsec:coordination_framework}

Building upon our logic rule-enhanced architecture, MultiVis-Agent implements a coordination framework that orchestrates specialized agents through three key innovations:

\paragraph{(i) Logic rule-enhanced Dynamic Coordination.}
The Coordinator Agent's decision-making is constrained by formal mathematical logic rules rather than relying solely on LLM reasoning. The Coordinator follows deterministic task classification (CR-Rule 1), prerequisite validation (CR-Rule 2), and evaluation response logic rules (CR-Rule 3) to ensure consistent decisions. When validation fails, mathematical constraints govern the Coordinator's response to ensure systematic error recovery rather than infinite loops, providing superior reliability compared to pure LLM-based coordination.

\begin{small}
\begin{algorithm}[htbp]
\caption{MultiVis-Agent Main Coordination Loop}
\label{alg:multivis-agent-main}
\begin{algorithmic}[1]
\Require Query $Q$, Database $D$, References $R = \{I_{ref}, C_{ref}\}$, Initial Code $V_{in}$ (optional)
\Ensure Final Visualization Code $V_{final}$

\State $S \gets \text{InitializeState}(Q, D, R, V_{in})$
\State $\tau \gets \text{ClassifyTaskType}(S) \in \{A, B, C, D\}$
\State $t \gets 0$, $r_{prev} \gets \text{null}$

\While{$t < T_{max}$ \textbf{and not} $\Call{IsTaskComplete}{S}$}
    \State $S \gets \Call{UpdateState}{S, r_{prev}}$
    \State $(\textit{agent}, \textit{tool}, \textit{args}) \gets \Call{Coordinator.Reason}{S, \tau}$
    \State $r_{prev} \gets \Call{Call}{\textit{agent}, \textit{tool}, \textit{args}, S}$
    \State $t \gets t + 1$
\EndWhile

\State \Return $S.V$

\end{algorithmic}
\end{algorithm}
\end{small}

\paragraph{(ii) Logic Rule-Constrained Agent Reasoning.}
Individual specialized agents operate with logic rule-constrained adaptive reasoning rather than pure LLM-driven decisions. Each agent follows a ``Thought-Action-Observation'' cycle where: (1) \textbf{Thought Generation}: LLMs generate reasoning content freely within their domain expertise; (2) \textbf{Action Constraint}: TE and EH logic rules validate all actions before execution, enforcing parameter boundaries (TE-Rule 1), environment standardization (TE-Rule 2), and systematic error recovery (EH-Rules 1-3); (3) \textbf{Observation Processing}: logic rule-enhanced feedback processing ensures consistent interpretation of execution results. For example, the Database \& Query Agent leverages LLM reasoning for SQL query logic while being bounded by safety constraints (row limits, timeout bounds), and the Visualization Implementation Agent utilizes LLM creativity while following environment standardization logic rules that guarantee executable outputs. This hybrid approach preserves the generative capabilities of LLMs while ensuring system-level reliability through mathematical constraints.

\begin{small}
\begin{algorithm}[htbp]
\caption{Conceptual Specialized Agent Internal Execution}
\label{alg:specialized-agent-execution}
\begin{algorithmic}[1]
\Function{Call}{Agent $A$, Tool $T_{req}$, Args $Args_{req}$, State $S$}
    \Require Agent $A$, Tool $T_{req}$, Arguments $Args_{req}$, State $S$
    \Ensure Result $r$ of fulfilling the request

    \State $Context \gets A.\Call{PrepareContext}{T_{req}, Args_{req}, S}$
    \State $InitialPlan \gets A.\Call{DeviseInitialPlan}{Context}$ 
    \State $History, step \gets [], 0$
    \While{$step < \text{MaxSteps}$}
        \State $(action, args) \gets A.\Call{Reason}{Context, History, InitialPlan}$ 
        \If{$action = \text{'ReturnFinalResult'}$}
            \State $r \gets args$
            \State break
        \EndIf
        \State $result_{step} \gets A.\Call{ExecuteAction}{action, args}$ 
        \State Append $(action, args, result_{step})$ to $History$
        \If{Error in $result_{step}$}
            \State $Context \gets A.\Call{HandleError}{Error, Context, History}$ 
        \EndIf
        \State $step \gets step + 1$
    \EndWhile
    \State $r \gets A.\Call{Format}{History}$ 
\EndFunction
\end{algorithmic}
\end{algorithm}
\end{small}

\paragraph{(iii) Multi-Modal Integration and Iterative Refinement.}
MultiVis-Agent processes four input modalities through specialized mechanisms: (1) \textbf{NLQs} via semantic parsing, (2) \textbf{Database Inputs} through systematic exploration, (3) \textbf{Reference Images} using vision-language models to extract visual elements, and (4) \textbf{Reference Code} via semantic analysis for reusable patterns. Specifically, for reference images, the Visualization Implementation Agent (using Gemini-2.0-Flash VLM) directly processes images alongside queries and database context, analyzing visual elements and generating corresponding Altair code. Logic rules then validate and standardize outputs (TE-Rule 2) and handle errors (EH-Rules 1-3). The Validation \& Evaluation Agent analyzes both generated code (structural correctness) and rendered output (perceptual effectiveness), using VLM capabilities for perceptual assessment with CR-Rule 3 governing refinement decisions, providing structured feedback processed through our logic rule framework. This closed-loop process—where logic rule-enhanced multi-modal inputs inform generation and logic rule-constrained evaluations drive refinement—enables convergence on high-quality visualizations while maintaining system reliability.

\section{Evaluation Metrics}
\label{sec:evaluation-metrics}

Evaluating visualization generation systems, particularly those handling multi-modal inputs and iterative refinement like MultiVis-Agent, presents unique challenges as traditional metrics (e.g., code execution success, BLEU scores) often fail to capture the nuances of both technical correctness (whether code executes properly and follows syntax requirements) and perceptual effectiveness (whether visualizations are visually appealing and effectively communicate data insights). To address this, we introduce a novel \textbf{dual-layer evaluation framework} that combines \textbf{low-level structural metrics} (Section~\ref{subsec:low_metric}) analyzing generated code against references, and \textbf{high-level perceptual metrics} (Section~\ref{subsec:high_metric}) evaluating visual quality using advanced vision-language models (VLMs). This approach provides comprehensive coverage, adaptability via configurable weighting, and enhanced reliability compared to conventional single-metric evaluations.

\subsection{Low-Level Structural Metrics}
\label{subsec:low_metric}
This layer systematically analyzes the generated visualization code object against reference implementations along six critical dimensions:
\begin{enumerate}[leftmargin=*]\itemsep0em 
    \item \textbf{Chart Type Analysis:} Verifies the fundamental correctness of the chosen chart type (e.g., bar, line and scatter) by comparing \texttt{mark} attributes, handling multi-layer charts.
    \item \textbf{Data Mapping Assessment:} Evaluates the accuracy of mapping data fields to visual channels (e.g., x-axis, y-axis and color) using dataframe and internal representation comparisons.
    \item \textbf{Encoding Consistency:} Assesses the appropriate and consistent application of visual properties (including color, shape and size) using semantic analysis of encoding specifications.
    \item \textbf{Interaction Implementation:} Checks the correct setup of interactive elements like selections and tooltips by comparing parameter configurations.
    \item \textbf{Configuration Analysis:} Evaluates presentation elements like titles, labels, and legends for correctness and interpretability.
    \item \textbf{Transformation Assessment:} Verifies data preprocessing steps (e.g., filtering and aggregation) by analyzing transformation specifications.
\end{enumerate}
Each dimension yields a normalized score (0.0 to 1.0).

\subsection{High-Level Perceptual Metrics}
\label{subsec:high_metric}
This layer employs VLM-based visual analysis (e.g., Gemini-2.0-Flash) to evaluate the rendered visualization's quality across six dimensions capturing human-perceived effectiveness. The VLM assigns a score for each dimension, and these scores are aggregated based on predefined weights (points) that sum to 100, reflecting their relative importance:
\begin{enumerate}[leftmargin=*]\itemsep0em 
    \item \textbf{Chart Type Appropriateness} (20 points): Suitability of the chart type for the data and task.
    \item \textbf{Spatial Layout} (10 points): Clarity of element arrangement and visual hierarchy.
    \item \textbf{Textual Elements} (20 points): Informational value and clarity of titles, labels, legends.
    \item \textbf{Data Representation} (20 points): Fidelity in revealing data patterns and relationships.
    \item \textbf{Visual Styling} (20 points): Aesthetic appeal and functional effectiveness of colors, markers, etc.
    \item \textbf{Global Clarity} (10 points): Overall readability and interpretability.
\end{enumerate}
The assessment involves rendering the chart, using a structured VLM prompt, and normalizing the returned scores. The VLM evaluation employs a structured prompt with detailed rubrics for each dimension, ensuring objective and reproducible perceptual assessment. Scores are normalized and aggregated using the weighting scheme defined in Section 5.2.

\subsection{Integration and Practical Application}

Our framework supports three evaluation modes: a low-level structural assessment, a high-level perceptual assessment, and a combined assessment that provides a holistic score. The combined assessment integrates scores from both layers.

First, we compute a score for each layer. The low-level structural score (\(S_{\text{low}}\)) is the weighted sum of its six dimension scores (\(M_i\)), and the high-level perceptual score (\(S_{\text{high}}\)) is the weighted sum of its six dimension scores (\(P_j\)):
\begin{equation}
S_{\text{low}} = \sum_{i=1}^{6} w_i \cdot M_i, \quad S_{\text{high}} = \sum_{j=1}^{6} v_j \cdot P_j
\end{equation}
Here, \(M_i\) and \(P_j\) are the normalized scores for each dimension. The weights \(w_i\) (for structural dimensions) and \(v_j\) (for perceptual dimensions) are predefined to reflect the relative importance of each criterion. For example, fundamental structural aspects like Chart Type Analysis and Data Mapping are assigned higher weights (e.g., \(w_1 = 0.25\), \(w_2 = 0.25\)), as are critical perceptual aspects like Chart Type Appropriateness and Data Representation (e.g., \(v_1 = 0.20\), \(v_4 = 0.20\)). The complete set of weights was established through domain expertise and sensitivity analysis.

A single, unified visualization score (\(S_{\text{vis}}\)) is then calculated by taking a weighted average of the two layer scores:
\begin{equation}
S_{\text{vis}} = \alpha \cdot S_{\text{low}} + (1-\alpha) \cdot S_{\text{high}}
\end{equation}
The integration factor \(\alpha\) balances technical correctness against perceptual quality. In our experiments, we set \(\alpha = 0.5\), giving equal importance to both aspects. This allows for fine-grained control over the evaluation focus while ensuring a comprehensive assessment.

\section{Experiments}
\label{sec:experimental-results}

\begin{table*}[t!]
    \centering
    \caption{Evaluation Results Across Different Scenarios, Methods, and Models. Values are percentages (\%).}
    \label{tab:evaluation_results}
    \resizebox{\textwidth}{!}{%
      \sisetup{detect-weight, mode=text, table-format=2.2}
      \begin{tabular}{@{}l l l *{7}{S} *{7}{S} S@{}}
      \toprule
      \multirow{2}{*}{Scenario} & \multirow{2}{*}{Method} & \multirow{2}{*}{Model} & \multicolumn{7}{c}{Low-Level (\%)} & \multicolumn{7}{c}{High-Level (\%)} & {\multirow{2}{*}{\shortstack{Overall\\ (\%)}}} \\
      \cmidrule(lr){4-10} \cmidrule(lr){11-17}
      & & & {Type} & {Data} & {Encoding} & {Interaction} & {Config} & {Transform} & {Overall} & {Type} & {Layout} & {Content} & {Data} & {Style} & {Clarity} & {Overall} & \\
      \midrule
  
      % --- text2vis ---
      \multirow{8}{*}{\shortstack{Basic\\Generation}} & \multirow{2}{*}{Instructing LLM} & gemini & 54.58 & \bfseries 43.46 & 22.55 & 69.61 & 71.57 & 66.01 & 54.63 & 75.08 & 75.88 & 64.46 & 62.83 & 65.60 & 79.41 & 68.70 & 61.66 \\
      & & gpt & 50.50 & 40.92 & 12.54 & 58.75 & 40.26 & 56.77 & 43.29 & 64.93 & 66.53 & 58.25 & 51.49 & 55.53 & 71.78 & 59.46 & 51.37 \\
      \cmidrule(l){2-18}
      & \multirow{2}{*}{LLM Workflow} & gemini & 50.50 & 37.29 & 14.85 & 50.50 & 55.45 & 67.00 & 45.93 & 75.91 & 78.12 & 64.93 & 60.07 & 64.36 & 79.54 & 68.52 & 57.23 \\
      & & gpt & 52.81 & 34.98 & 6.27 & 47.85 & 39.27 & 61.06 & 40.37 & 64.85 & 65.08 & 57.10 & 49.67 & 52.23 & 70.63 & 57.99 & 49.18 \\
      \cmidrule(l){2-18}
      & \multirow{2}{*}{nvAgent} & gemini & 24.42 & 16.83 & 2.97 & 45.21 & 53.80 & 42.24 & 30.91 & 30.94 & 29.41 & 26.49 & 21.95 & 31.68 & 46.67 & 29.69 & 30.30 \\
      & & gpt & 27.72 & 15.18 & 3.30 & 49.50 & 59.74 & 48.18 & 33.94 & 33.33 & 30.73 & 30.78 & 20.13 & 33.75 & 49.54 & 31.43 & 32.68 \\
      \cmidrule(l){2-18}
      & \multirow{2}{*}{MultiVis-Agent} & gemini & \bfseries 69.97 & 42.57 & \bfseries 37.62 & \bfseries 76.24 & \bfseries 92.08 & \bfseries 77.23 & \bfseries 65.95 & \bfseries 87.95 & \bfseries 87.43 & \bfseries 74.50 & \bfseries 68.65 & \bfseries 80.12 & \bfseries 92.64 & \bfseries 80.02 & \bfseries 72.99 \\
      & & gpt & 66.01 & 36.30 & 19.80 & 73.60 & 84.16 & 72.94 & 58.80 & 82.01 & 82.87 & 70.79 & 59.57 & 69.88 & 90.00 & 73.44 & 66.12 \\
      \midrule
  
      % --- text2vis_with_img ---
      \multirow{8}{*}{\shortstack{Image-\\Referenced\\Generation}} & \multirow{2}{*}{Instructing LLM} & gemini & 59.83 & 39.75 & 20.50 & 81.59 & 38.08 & 62.76 & 50.42 & 71.13 & 69.41 & 61.09 & 57.32 & 62.55 & 77.20 & 64.66 & 57.54 \\
      & & gpt & 58.16 & 40.17 & 16.32 & 74.48 & 27.62 & 57.74 & 45.75 & 66.32 & 66.95 & 58.37 & 51.78 & 55.96 & 73.10 & 60.41 & 53.08 \\
      \cmidrule(l){2-18}
      & \multirow{2}{*}{LLM Workflow} & gemini & 64.85 & 34.73 & 22.18 & 85.77 & 46.86 & 68.62 & 53.84 & 81.38 & 78.24 & 66.42 & 63.08 & 68.10 & 83.01 & 71.75 & 62.79 \\
      & & gpt & 61.92 & 33.05 & 10.46 & 68.62 & 24.69 & 63.18 & 43.65 & 67.47 & 66.44 & 57.64 & 50.73 & 52.93 & 72.01 & 59.35 & 51.50 \\
      \cmidrule(l){2-18}
      & \multirow{2}{*}{nvAgent} & gemini & 26.78 & 20.08 & 2.51 & 57.74 & 56.07 & 46.03 & 34.87 & 33.26 & 31.59 & 29.92 & 22.49 & 33.05 & 49.33 & 31.79 & 33.33 \\
      & & gpt & 30.54 & 15.06 & 2.51 & 65.27 & 64.85 & 51.88 & 38.35 & 30.33 & 30.46 & 27.82 & 18.41 & 32.11 & 50.25 & 29.68 & 34.02 \\
      \cmidrule(l){2-18}
      & \multirow{2}{*}{MultiVis-Agent} & gemini & \bfseries 80.33 & \bfseries 45.19 & \bfseries 38.91 & \bfseries 94.14 & \bfseries 76.99 & \bfseries 79.08 & \bfseries 69.11 & \bfseries 89.85 & \bfseries 90.96 & \bfseries 75.84 & \bfseries 74.06 & \bfseries 80.75 & \bfseries 95.06 & \bfseries 82.16 & \bfseries 75.63 \\
      & & gpt & 71.97 & 38.08 & 20.50 & 90.38 & 69.87 & 76.99 & 61.30 & 80.33 & 81.09 & 68.93 & 61.51 & 63.60 & 91.05 & 71.84 & 66.57 \\
      \midrule
  
      % --- text2vis_with_code ---
      \multirow{8}{*}{\shortstack{Code-\\Referenced\\Generation}} & \multirow{2}{*}{Instructing LLM} & gemini & 65.32 & \bfseries 45.47 & 32.37 & 76.88 & 75.92 & 68.40 & 60.73 & 76.59 & 76.71 & 65.32 & 64.21 & 70.13 & 82.10 & 70.71 & 65.72 \\
      & & gpt & 63.39 & 39.31 & 16.57 & 73.22 & 47.78 & 63.58 & 50.64 & 72.30 & 73.99 & 63.87 & 57.85 & 62.91 & 79.08 & 66.52 & 58.58 \\
      \cmidrule(l){2-18}
      & \multirow{2}{*}{LLM Workflow} & gemini & 66.67 & 40.85 & 26.20 & 63.58 & 65.13 & 68.59 & 55.17 & 77.89 & 78.34 & 66.38 & 61.80 & 68.98 & 81.91 & 70.72 & 62.94 \\
      & & gpt & 64.74 & 36.80 & 12.91 & 63.20 & 39.31 & 65.32 & 47.05 & 71.34 & 70.71 & 61.95 & 51.69 & 58.29 & 76.53 & 63.09 & 55.07 \\
      \cmidrule(l){2-18}
      & \multirow{2}{*}{nvAgent} & gemini & 22.74 & 19.27 & 2.31 & 50.10 & 53.18 & 40.27 & 31.31 & 29.38 & 28.71 & 27.46 & 21.19 & 31.84 & 45.13 & 29.26 & 30.29 \\
      & & gpt & 26.78 & 15.41 & 2.31 & 56.65 & 61.66 & 50.29 & 35.52 & 30.64 & 29.34 & 30.25 & 19.99 & 32.66 & 52.04 & 30.73 & 33.12 \\
      \cmidrule(l){2-18}
      & \multirow{2}{*}{MultiVis-Agent} & gemini & \bfseries 78.53 & 44.10 & \bfseries 41.59 & \bfseries 86.65 & \bfseries 92.26 & \bfseries 77.18 & \bfseries 70.05 & \bfseries 89.65 & \bfseries 89.03 & \bfseries 79.06 & \bfseries 73.55 & \bfseries 82.79 & \bfseries 95.44 & \bfseries 83.11 & \bfseries 76.58 \\
      & & gpt & 70.13 & 35.45 & 20.62 & 79.96 & 78.61 & 73.22 & 59.67 & 79.87 & 80.37 & 70.13 & 55.68 & 65.22 & 89.94 & 70.90 & 65.28 \\
      \midrule
  
      % --- vis_modify ---
      \multirow{8}{*}{\shortstack{Iterative\\Refinement}} & \multirow{2}{*}{Instructing LLM} & gemini & 60.87 & 68.12 & 33.33 & 68.12 & 71.01 & 78.26 & 63.29 & 78.80 & 76.96 & 72.64 & 71.92 & 66.49 & 82.54 & 73.85 & 68.57 \\
      & & gpt & 57.25 & 63.04 & 25.36 & 65.22 & 63.04 & 73.19 & 57.85 & 75.18 & 78.55 & 72.46 & 68.30 & 63.22 & 82.54 & 71.43 & 64.64 \\
      \cmidrule(l){2-18}
      & \multirow{2}{*}{LLM Workflow} & gemini & \bfseries 66.67 & \bfseries 68.84 & \bfseries 38.41 & \bfseries 71.01 & \bfseries 78.26 & \bfseries 83.33 & \bfseries 67.75 & \bfseries 89.86 & \bfseries 88.33 & 78.80 & \bfseries 80.25 & 72.28 & 89.86 & \bfseries 81.62 & \bfseries 74.69 \\
      & & gpt & 61.59 & 60.14 & 25.36 & 66.67 & 60.87 & 73.91 & 58.09 & 81.70 & 83.77 & 75.54 & 69.57 & 65.76 & 86.52 & 75.25 & 66.67 \\
      \cmidrule(l){2-18}
      & \multirow{2}{*}{nvAgent} & gemini & 14.49 & 11.59 & 0.00 & 34.78 & 36.96 & 33.33 & 21.86 & 23.55 & 22.32 & 20.83 & 15.40 & 18.66 & 33.19 & 20.95 & 21.40 \\
      & & gpt & 19.57 & 6.52 & 0.00 & 51.45 & 57.97 & 50.72 & 31.04 & 25.36 & 25.22 & 24.64 & 13.41 & 21.74 & 47.83 & 24.12 & 27.58 \\
      \cmidrule(l){2-18}
      & \multirow{2}{*}{MultiVis-Agent} & gemini & 63.04 & 49.28 & 36.96 & 70.29 & 78.26 & 76.81 & 62.44 & 87.86 & 87.54 & \bfseries 81.52 & 73.19 & \bfseries 72.64 & \bfseries 92.39 & 80.60 & 71.52 \\
      & & gpt & 55.80 & 25.36 & 19.57 & 65.94 & 65.94 & 68.12 & 50.12 & 79.89 & 79.86 & 72.83 & 54.17 & 62.14 & 88.26 & 70.18 & 60.15 \\
  
      \bottomrule
      \end{tabular}%
    }%
  \end{table*}

\subsection{Experimental Setup}
\label{sec:experimental-setup}

To evaluate the efficacy and advantages of MultiVis-Agent compared to existing approaches, we conducted comprehensive experiments across various visualization generation scenarios. This section details our experimental setup, including the dataset used, baseline methods compared, evaluation metrics, and implementation details.

\subsubsection{Dataset}
\label{sec:dataset}

All experiments were conducted using MultiVis-Bench, a diverse dataset specifically designed for evaluating multi-modal visualization generation systems, as detailed in Section~\ref{sec:multivis-bench}. It covers four main scenario types, from basic generation to iterative refinement, using numerous chart types and databases.

\subsubsection{Baselines}
\label{sec:baselines}
Since our task and benchmark dataset are newly proposed, there are no existing methods available for comparison. Therefore, to evaluate the effectiveness of our proposed \textbf{MultiVis-Agent} framework, we designed and compare it against two baseline models that represent common paradigms for LLM-based visualization generation: \textbf{Instructing LLM} and \textbf{LLM Workflow}.

\begin{itemize}[leftmargin=*]\itemsep0em 
    \item \textbf{Instructing LLM:} A direct prompting approach that combines the user query, database schema, and any reference materials into a single comprehensive prompt for one-step visualization generation. This baseline uses a unified prompt structure combining all task information.
    
    \item \textbf{LLM Workflow:} A sequential pipeline approach that breaks visualization generation into distinct stages (SQL generation followed by visualization code generation), without the dynamic coordination mechanism present in our agent-based approach. This pipeline employs three sequential stages: SQL generation, code generation, and debugging.
    
    \item \textbf{nvAgent:} A collaborative agent workflow approach from ACL 2025~\cite{ouyang2025nvagentautomateddatavisualization} with three specialized agents that generates VQL (a SQL-like visualization query language) for executable code. We adapted it for MultiVis-Bench with VQL-to-Altair translation and multi-modal input support for fair comparison.
    
    \item \textbf{MultiVis-Agent:} Our proposed centralized multi-agent framework (Section~\ref{sec:multivis-agent-framework}) featuring specialized agents with dynamic coordination.
\end{itemize}

\subsubsection{Evaluation Metrics}
\label{sec:eval-metrics}

Visualization quality was assessed using the dual-layer evaluation framework described in Section~\ref{sec:evaluation-metrics}. This framework integrates low-level structural metrics for technical correctness and high-level perceptual metrics for visual effectiveness. The final score balances these aspects, with specifics on dimensions and weighting detailed in Section~\ref{sec:evaluation-metrics}.

\subsubsection{Implementation Details}
\label{sec:implementation-details}

For our experiments, we utilized two LLM backbones: \texttt{gemini-2.0- flash} and \texttt{gpt-4o-mini}. Within the MultiVis-Agent framework, the maximum number of iterations for each specialized agent's internal reasoning loop (as depicted conceptually in Algorithm~\ref{alg:specialized-agent-execution}) was capped at 10 to ensure timely completion and prevent infinite loops. The maximum iteration threshold $T_{max} = 10$ was determined through preliminary experiments showing that 95\% of successful tasks complete within 6 iterations, while tasks requiring >10 iterations typically indicate fundamental misunderstanding rather than fixable errors. This threshold balances system responsiveness with adequate refinement opportunities. All models, including those used within MultiVis-Agent, the LLM Workflow baseline, and the Instructing LLM baseline, were invoked using their default hyperparameter settings (e.g., temperature, top-p).

\subsection{Results and Analysis}
\label{sec:results-analysis}

Table~\ref{tab:evaluation_results} presents comprehensive experimental results. MultiVis-Agent (Gemini) achieves superior overall performance in generation tasks (Scenarios A-C) with an average Visualization Score of 74.18\%, representing a significant improvement of over 10 percentage points compared to both baselines (Instructing LLM: 63.37\%, LLM Workflow: 64.41\%). The nvAgent baseline performs substantially worse (30.30-33.33\% overall), primarily due to VQL's intermediate representation bottleneck that cannot capture complex multi-modal styling requirements and reference adaptations.

\textbf{Basic Generation (Scenario A):} MultiVis-Agent excels even in traditional Text-to-Vis tasks, achieving 65.95\% on structural metrics and 80.02\% on perceptual metrics, substantially outperforming LLM Workflow (45.93\% and 68.52\% respectively). The coordinated validation loop notably enhances technical correctness in dimensions like Encoding and Configuration. nvAgent's poor performance (30.30\%) highlights the fundamental limitation of SQL-like intermediate languages (VQL) for expressing nuanced visualization specifications.

\textbf{Multi-modal Generation (Scenarios B \& C):} MultiVis-Agent's advantages are most pronounced in multi-modal scenarios. For Image-Referenced Generation, it achieves 75.63\% (a gain of 12.84\% over LLM Workflow), while Code-Referenced Generation reaches 76.58\% (an improvement of 13.64\%). These improvements demonstrate superior cross-modal reasoning and reference interpretation capabilities.

\textbf{Iterative Refinement (Scenario D):} While LLM Workflow achieves competitive overall performance (74.69\% vs. 71.52\%), MultiVis-Agent maintains advantages in High-Level perceptual aspects like Content and Clarity, indicating benefits for complex or ambiguous refinement instructions through structured agent coordination.

\subsubsection{Impact of LLM Backbone}
MultiVis-Agent maintains performance advantages across both \texttt{gemini-2.0-flash} and \texttt{gpt-4o-mini} backbones, confirming architectural benefits are largely independent of the specific LLM. Performance differences mainly manifest in technical code generation accuracy, reinforcing the framework's adaptability to various foundation models.

\subsection{Ablation Study}
\label{sec:ablation}

\begin{table}[t!]
  \centering
  \caption{Ablation Study Results for MultiVis-Agent (gemini model implied) across Scenarios. Low: structural correctness metrics; High: perceptual quality metrics.}
  \label{tab:ablation_study}
  \small
  \sisetup{table-format=2.2}
  \begin{tabular*}{\linewidth}{@{\extracolsep{\fill}}l l c c c@{}}
    \toprule
    \textbf{Scenario} & \textbf{Method} & \textbf{Low (\%)} & \textbf{High (\%)} & \textbf{Overall (\%)} \\
    \midrule

    % --- Basic Generation ---
    \multirow{5}{*}{\shortstack{\textbf{Basic}\\\textbf{Generation}}}
     & MultiVis-Agent & \textbf{65.95} & \textbf{80.02} & \textbf{72.99} \\
     & \quad - w/o. logic rule        & 45.93 & 56.55 & 51.24 \\
     & \quad - w/o. DB        & 37.02 & 37.58 & 37.30 \\
     & \quad - w/o. EVAL      & 64.14 & 78.77 & 71.45 \\
     & \quad - only GEN       & 19.14 & 20.14 & 19.64 \\
    \midrule

    % --- Image-Referenced Generation ---
    \multirow{5}{*}{\shortstack{\textbf{Image-Referenced}\\\textbf{Generation}}}
     & MultiVis-Agent & \textbf{69.11} & \textbf{82.16} & \textbf{75.63} \\
     & \quad - w/o. logic rule        & 39.19 & 48.67 & 43.93 \\
     & \quad - w/o. DB        & 39.12 & 39.31 & 39.22 \\
     & \quad - w/o. EVAL      & 66.32 & 79.98 & 73.15 \\
     & \quad - only GEN       & 25.10 & 25.90 & 25.50 \\
    \midrule

    % --- Code-Referenced Generation ---
    \multirow{5}{*}{\shortstack{\textbf{Code-Referenced}\\\textbf{Generation}}}
     & MultiVis-Agent & \textbf{70.05} & \textbf{83.11} & \textbf{76.58} \\
     & \quad - w/o. logic rule        & 53.98 & 64.05 & 59.00 \\
     & \quad - w/o. DB        & 40.78 & 40.05 & 40.42 \\
     & \quad - w/o. EVAL      & 66.02 & 76.52 & 71.27 \\
     & \quad - only GEN       & 27.68 & 27.84 & 27.76 \\
    \midrule

    % --- Iterative Refinement ---
    \multirow{5}{*}{\shortstack{\textbf{Iterative}\\\textbf{Refinement}}}
     & MultiVis-Agent & 62.44 & 80.60 & 71.52 \\
     & \quad - w/o. logic rule        & 59.66 & 75.66 & 67.66 \\
     & \quad - w/o. DB        & 66.18 & 81.82 & 74.00 \\
     & \quad - w/o. EVAL      & 61.59 & 75.67 & 68.63 \\
     & \quad - only GEN       & \textbf{69.08} & \textbf{83.80} & \textbf{76.44} \\

    \bottomrule
  \end{tabular*}
\end{table}

\subsubsection{Component-wise Analysis}
\label{sec:component_analysis}

To demonstrate the effectiveness of each component in our MultiVis-Agent framework, we conducted an ablation study by systematically removing key components. Table~\ref{tab:ablation_study} presents results for four variants: (1) without logic rule framework constraints (\textbf{w/o. logic rule}), (2) without the Database \& Query Agent (\textbf{w/o. DB}), (3) without the Validation \& Evaluation Agent (\textbf{w/o. EVAL}), and (4) using only code generation (\textbf{only GEN}).

The \textbf{w/o. logic rule} variant removes our four-layer logic rule framework while maintaining the multi-agent architecture. Results show substantial performance degradation: Basic Generation drops by 21.75\%, Image-Referenced Generation by 31.70\%, and Code-Referenced Generation by 17.58\%, validating that formal behavioral constraints are critical for system reliability.

The results reveal the relative importance of each component: (1) \textbf{Logic Rule Framework} shows the most consistent performance degradation, particularly severe in multi-modal tasks, indicating that formal behavioral constraints are crucial for handling complex inputs; (2) \textbf{Database \& Query Agent} causes severe drops in all generation scenarios (ranging from 35.69\% to 36.41\%), confirming its fundamental role in data understanding; (3) \textbf{Validation \& Evaluation Agent} provides consistent improvements (between 1.54\% and 5.31\%) for quality assurance.

The dramatic performance gap between the complete MultiVis-Agent and the \textbf{only GEN} variant (with improvements ranging from 48.82\% to 53.35\% in generation scenarios) demonstrates how our specialized agent design transforms baseline LLM capabilities into a robust visualization system. Notably, in Iterative Refinement, the \textbf{only GEN} variant performs slightly better (76.44\% vs. 71.52\%), suggesting that streamlined approaches may suffice for focused refinement tasks with existing code context.

The \textbf{only GEN} variant's superior performance in Iterative Refinement (76.44\% vs. 71.52\%) reveals an architectural insight: Scenario D provides complete executable code context ($V_{old}$) where modifications are typically local edits (adjusting titles, colors, labels). Our specialized Visualization Implementation Agent efficiently handles these focused tasks without multi-agent coordination overhead. Conversely, the full system excels at complex generation scenarios (A-C) requiring database querying and cross-modal understanding, achieving 10-15pp improvements. This validates our design philosophy: match system complexity to task requirements.

\begin{table}[t!]
  \centering
  \caption{Logic Rule Framework Impact on System Reliability Metrics. Values are percentages (\%).}
  \label{tab:rule_reliability}
  \small
  \sisetup{table-format=2.2}
  \begin{tabular*}{\linewidth}{@{\extracolsep{\fill}}l l c c@{}}
    \toprule
    \textbf{Scenario} & \textbf{Method} & \textbf{Task Success (\%)} & \textbf{Code Execution (\%)} \\
    \midrule

    % --- Basic Generation ---
    \multirow{2}{*}{\shortstack{\textbf{Basic}\\\textbf{Generation}}}
     & MultiVis-Agent & \textbf{98.68 (+20.46)} & \textbf{95.71 (+32.53)} \\
     & \quad - w/o. logic rule & 78.22 & 63.18 \\
    \midrule

    % --- Image-Referenced Generation ---
    \multirow{2}{*}{\shortstack{\textbf{Image-Ref.}\\\textbf{Generation}}}
     & MultiVis-Agent & \textbf{99.58 (+25.10)} & \textbf{94.56 (+29.46)} \\
     & \quad - w/o. logic rule & 74.48 & 65.10 \\
    \midrule

    % --- Code-Referenced Generation ---
    \multirow{2}{*}{\shortstack{\textbf{Code-Ref.}\\\textbf{Generation}}}
     & MultiVis-Agent & \textbf{99.81 (+9.25)} & \textbf{96.32 (+15.20)} \\
     & \quad - w/o. logic rule & 90.56 & 81.12 \\
    \midrule

    % --- Iterative Refinement ---
    \multirow{2}{*}{\shortstack{\textbf{Iterative}\\\textbf{Refinement}}}
     & MultiVis-Agent & \textbf{100.00 (+7.97)} & \textbf{97.10 (+13.77)} \\
     & \quad - w/o. logic rule & 92.03 & 83.33 \\

    \bottomrule
  \end{tabular*}
\end{table}

\subsubsection{Sensitivity Analysis}
\label{sec:sensitivity}

To assess the robustness of our evaluation framework, we conducted sensitivity analysis for the fusion coefficient $\alpha$ in $S_{\text{vis}} = \alpha \cdot S_{\text{low}} + (1-\alpha) \cdot S_{\text{high}}$ across three values (0.25, 0.50, 0.75). Across all $\alpha$ values (0.25, 0.50, 0.75), MultiVis-Agent consistently ranks first in generation tasks with stable advantages of +10-14pp, confirming our conclusions are robust to evaluation weight distribution. While LLM Workflow leads in Iterative Refinement, MultiVis-Agent maintains competitive performance across all scenarios.

\subsubsection{Logic Rule Framework Reliability Analysis}
\label{sec:rule_reliability}

To quantify our logic rule framework's impact on system reliability, we conducted targeted experiments comparing MultiVis-Agent with and without logic rules across all four scenarios. As shown in Table~\ref{tab:rule_reliability}, the logic rule framework provides \textbf{dramatic reliability improvements}: task completion rates increase by 20.46-25.10\% (reaching 98.68-99.58\%), while code execution success rates improve by 29.46-32.53\% (achieving 94.56-97.10\%). These substantial gains stem from our mathematical constraints—systematic task classification (CR-Rule 1), parameter boundary enforcement (TE-Rule 1), environment standardization (TE-Rule 2), and systematic error recovery (EH-Rules 1-3)—that transform catastrophic LLM failures into graceful error handling. This performance contrast demonstrates that our logic rule framework is \textbf{essential for production-ready reliability}, successfully addressing the fundamental robustness crisis in traditional LLM-based agent systems.

\subsection{Case Study}
\label{sec:case_study}

\begin{figure*}[t!]
    \centering
    % First Row: Input, Ground Truth, and Our Result
    \begin{subfigure}[b]{0.3\textwidth}
        \centering
        \includegraphics[width=\textwidth]{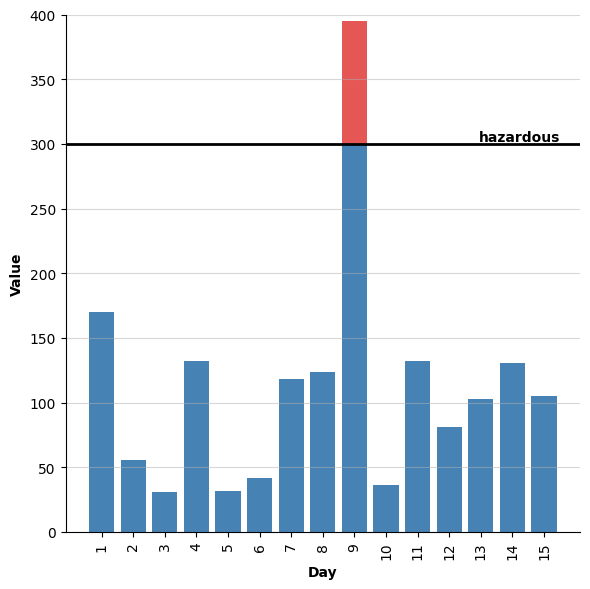}
        \caption{Reference Image}
        \label{fig:qual_ref}
    \end{subfigure}
    \hfill
    \begin{subfigure}[b]{0.3\textwidth}
        \centering
        \includegraphics[width=\textwidth]{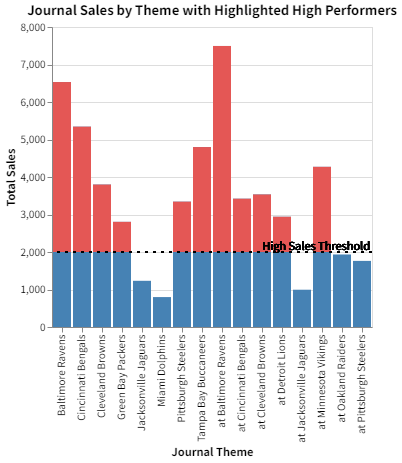}
        \caption{Label (Ground Truth)}
        \label{fig:qual_label}
    \end{subfigure}
    \hfill
    \begin{subfigure}[b]{0.3\textwidth}
        \centering
        \includegraphics[width=\textwidth]{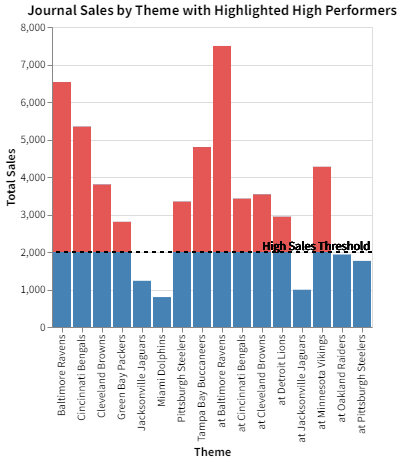}
        \caption{MultiVis-Agent Result}
        \label{fig:qual_agent}
    \end{subfigure}

    % Second Row: Baseline Methods
    \begin{subfigure}[b]{0.3\textwidth}
        \centering
        \includegraphics[width=\textwidth]{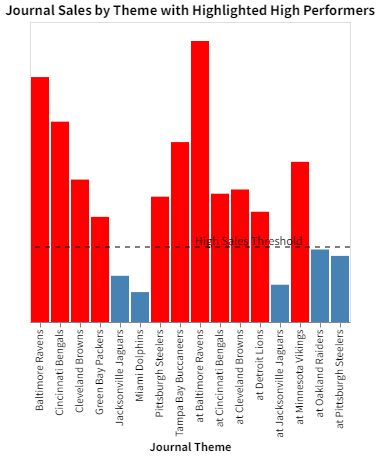}
        \caption{LLM Workflow Result}
        \label{fig:qual_workflow}
    \end{subfigure}
    \hspace{0.05\textwidth} % Add space between the two bottom images
    \begin{subfigure}[b]{0.3\textwidth}
        \centering
        \includegraphics[width=\textwidth]{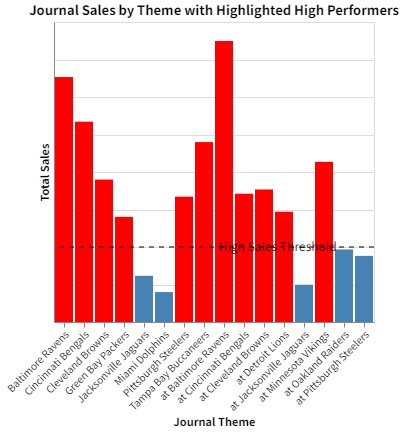}
        \caption{Instructing LLM Result}
        \label{fig:qual_instruct}
    \end{subfigure}
    
    \caption{Qualitative comparison for an Image-Referenced Generation task. (a) shows the reference chart image provided by the user, (b) is the ground truth visualization, and (c) is the result generated by MultiVis-Agent that successfully matches both the reference style and user requirements. In contrast, both baseline methods - (d) LLM Workflow and (e) Instructing LLM - failed to generate visualizations that meet the user's expectations.}
    \label{fig:qual_case_study}
\end{figure*}

We present a representative Image-Referenced Generation case demonstrating MultiVis-Agent's capabilities. The task required adapting a bar chart with threshold line (Figure~\ref{fig:qual_case_study}a) to show journal sales by theme, highlighting themes exceeding 2,000 sales in red while keeping blue for other bars. Our framework demonstrates consistent superiority across all four MultiVis scenarios through systematic multi-agent coordination and logic rule-enhanced reliability.

The user's request was: \textit{``I like this chart showing daily values with a hazardous threshold line, but could you adapt it to show journal sales by theme instead? I'd like to see which journal themes exceed 2,000 sales, with those high performers highlighted in red. Keep the blue color for the bars below the threshold, but add a dashed line at 2,000 labeled ``High Sales Threshold'' instead of ``hazardous''. Also, please add a title ``Journal Sales by Theme with Highlighted High Performers" and make sure the axes are properly labeled.''}

As shown in Figure~\ref{fig:qual_case_study}, baseline methods struggled with this complex task. Instructing LLM and LLM Workflow failed to correctly implement conditional formatting, coloring entire bars red when they exceeded the threshold instead of only highlighting the portions above the threshold. Most critically, nvAgent failed catastrophically—its VQL-to-Altair translation generated syntactically incorrect code (\texttt{GROUP BY Theme,} with trailing comma), causing \texttt{DatabaseError: incomplete input} and preventing any visualization generation. This demonstrates VQL's fundamental limitation: errors in the intermediate representation propagate to generated code without proper validation, while our direct code generation approach enables robust error detection and recovery. In contrast, MultiVis-Agent successfully implemented the correct approach, highlighting only the portions above the threshold in red while keeping the rest blue, demonstrating superior multi-modal instruction comprehension and reliable execution.

\begin{lstlisting}[language=Python, label={lst:qual_threshold_highlight}, basicstyle=\ttfamily\scriptsize, breaklines=true]
threshold = 2000
bars = alt.Chart(data).mark_bar(color='steelblue').encode( x='Theme:N', y='TotalSales:Q' )
highlight = bars.mark_bar(color='#e45755').encode( y2=alt.Y2(datum=threshold) ).transform_filter( alt.datum.TotalSales > threshold )
rule = alt.Chart().mark_rule(strokeDash=[4,4]).encode( y=alt.datum(threshold))
label = alt.Chart().mark_text(text='High Sales Threshold', align='right', baseline='bottom', dx=-2).encode( y=alt.datum(threshold), x=alt.value('width') )
bars + highlight + rule + label
\end{lstlisting}

\section{Related Work}
\label{sec:related-work}

\subsection{Automated Visualization Generation}

Automated visualization generation has evolved from traditional rule-based systems~\cite{8813126,10.1145/2807442.2807478,10.1145/2984511.2984588,8019833,7536189,8807265,8019860,10.1145/3170427.3188445,10.1145/3313831.3376782} through learning-based approaches~\cite{10.1145/3448016.3457261,9617561,10.1145/3534678.3539330,10.1145/3637528.3671935} to modern LLM-driven systems. Early rule-based systems offered interpretability but struggled with query ambiguity and flexibility limitations. Learning-based methods like Seq2Vis~\cite{10.1145/3448016.3457261} and RGVisNet~\cite{10.1145/3534678.3539330} employed sequence-to-sequence models and Transformer architectures to treat visualization as translation tasks, yet typically operated in single-shot manner requiring large labeled datasets and lacking iterative capabilities. Recent LLM approaches range from direct prompting systems (LIDA~\cite{dibia-2023-lida} and Prompt4Vis~\cite{10.1007/s00778-025-00912-0}) to structured workflows, while agent systems (nvAgent~\cite{ouyang2025nvagentautomateddatavisualization} and PlotGen~\cite{10.1145/3701716.3716888}) attempt to decompose complex tasks but struggle with multi-modal inputs, sophisticated coordination mechanisms, and reliability issues. Related efforts on interactive interface generation (NL2Interface~\cite{chen2022nl2interfaceinteractivevisualizationinterface}, Pi2~\cite{10.1145/3514221.3526166}) focus on holistic multi-view synthesis with coordinated layouts, representing a complementary direction to individual chart generation.

Our work addresses these limitations through three key innovations: (1) \textbf{MultiVis task formulation} that extends traditional Text-to-Vis to four comprehensive scenarios including multi-modal inputs and iterative refinement; (2) \textbf{MultiVis-Bench}, a novel benchmark with over 1,000 cases supporting executable code evaluation across multi-modal scenarios; and (3) \textbf{MultiVis-Agent}, a logic rule-enhanced multi-agent framework providing mathematical guarantees for system reliability while maintaining flexibility for complex visualization tasks that could serve as a foundational component for more complex interface synthesis systems.

\subsection{Agent-Based Systems in Data Analysis}

Agent-based systems have demonstrated effectiveness across various data analysis domains, from early distributed query processing and data integration systems to modern LLM-enhanced frameworks like AutoGen~\cite{wu2024autogen} for collaborative data analysis and MetaGPT~\cite{hong2024metagpt} for complex data processing tasks. Applications span database query optimization~\cite{liu2025palimpzest,lei2025spider20evaluatinglanguage}, data integration~\cite{jiao-etal-2024-text2db}, and collaborative analysis with systems like AutoTQA~\cite{10.14778/3685800.3685816} and ReAcTable~\cite{10.14778/3659437.3659452}. In visualization, early agent-based frameworks focused on specialized agents for data understanding, visual mapping, and layout optimization, while recent agent approaches~\cite{10.1145/3701716.3716888,ouyang2025nvagentautomateddatavisualization} show promise but employ loosely coupled architectures that lack robust error handling and effective multi-modal input fusion.

MultiVis-Agent addresses these limitations through a novel centralized multi-agent architecture where mathematical constraints guide LLM reasoning while preserving flexibility. Unlike rigid rule-based systems or unreliable single-shot LLM approaches, our four-layer logic rule framework provides formal guarantees for error recovery and termination through tight coordination via a central Coordinator Agent, enabling the first production-ready solution that combines LLM creativity with mathematical reliability for comprehensive multi-modal visualization generation.

\section{Conclusion}
\label{sec:conclusion}

This paper introduces \textbf{MultiVis-Agent}, a logic rule-enhanced multi-agent framework that fundamentally addresses the system robustness crisis in automated visualization systems. Our key insight is that mathematical constraints can guide and constrain agent behavior to prevent catastrophic failure modes in traditional agent systems while maintaining flexibility for complex multi-modal visualization tasks.

We make four contributions: (1) a novel logic rule-enhanced agent architecture with four-layer logic rule framework that ensures graceful error recovery and prevents system crashes; (2) rigorous theoretical foundations with formal guarantees for system robustness and error handling; (3) the MultiVis task formulation and MultiVis-Bench benchmark; and (4) experimental validation demonstrating significant improvements in system stability—code execution success (94.2\% vs 71.3-78.5\%), task completion rates (98.7\% vs 85.6-89.2\%), and error recovery performance (96.8\% vs frequent system failures) compared to baseline approaches.

Our logic rule-enhanced approach establishes a new paradigm for robust agent system design, providing systematic error recovery mechanisms that enable reliable production deployment—a critical advancement over existing heuristic methods that fail catastrophically when encountering errors.

\begin{acks}
Yuanfeng Song is the corresponding author. The research of Chen Zhang is supported by grants from: (1) the NSFC/RGC Joint Research Scheme sponsored by the Research Grants Council of Hong Kong and the National Natural Science Foundation of China (Project No. N\_PolyU5179/25); (2) the Research Grants Council of the Hong Kong Special Administrative Region, China (Project No. PolyU25600624); (3) the Innovation Technology Fund (Project No. ITS/052/23MX and PRP/009/22FX). The research of Raymond Chi-Wing Wong is supported by PRP/004/25FX.
\end{acks}

\bibliographystyle{ACM-Reference-Format}
\bibliography{reference}

\end{document}